\documentclass{article}

\PassOptionsToPackage{square,numbers,sort&compress}{natbib}

\usepackage[preprint]{neurips_2025}

\usepackage[utf8]{inputenc} 
\usepackage[T1]{fontenc}    
\usepackage{hyperref}       
\usepackage{url}            
\usepackage{booktabs}       
\usepackage{amsfonts}       
\usepackage{algorithm}      
\usepackage{algorithmic}      
\usepackage{nicefrac}       
\usepackage{microtype}      
\usepackage{xcolor}         

\usepackage{amsmath}
\usepackage{amssymb}
\usepackage{mathtools}
\usepackage{amsthm}
\usepackage{textgreek}
\usepackage{lgreek}
\usepackage{comment}

\usepackage[capitalize,noabbrev]{cleveref}

\usepackage{wrapfig}

\usepackage{microtype}
\usepackage{subcaption}
\usepackage{graphicx}
\usepackage{booktabs} 
\usepackage{multirow}
\usepackage{xspace}
\usepackage{bm}
\usepackage{epstopdf}
\usepackage{paralist}

\theoremstyle{plain}
\newtheorem{theorem}{Theorem}[section]

\newtheorem{lemma}[theorem]{Lemma}

\theoremstyle{definition}

\theoremstyle{remark}

\newcommand{\ind}{\perp\!\!\!\!\perp} 

\newcommand{\arce}{ARC-E\xspace}
\newcommand{\arcc}{ARC-C\xspace}
\newcommand{\hell}{Hellaswag\xspace}
\newcommand{\humaneval}{Human-eval\xspace}
\newcommand{\qmsum}{LongBench-qmsum\xspace}
\newcommand{\govreport}{LongBench-gov\_report\xspace}
\newcommand{\thegreektheta}{\begin{greek}j\end{greek}}
\newcommand{\topth}{Top-\thegreektheta\xspace}
\newcommand{\topk}{Top-k\xspace}
\newcommand{\calibrationSet}{\mathcal{C}}
\newcommand{\selids}{\mathcal{I}}
\newcommand{\notselids}{\Bar{\mathcal{I}}}
\newcommand{\seqlen}{n}
\newcommand{\headdim}{d}
\newcommand{\embdim}{D}
\newcommand{\numKept}{\Tilde{k}}
\newcommand{\attnMatPre}{\bm{A}}
\newcommand{\attnVecPre}{\bm{a}}
\newcommand{\attnVecSparsePre}{\Tilde{\attnVecPre}}
\newcommand{\attnMatPost}{\bm{S}}
\newcommand{\attnVecPost}{\bm{s}}
\newcommand{\attnVecSparsePost}{\Tilde{\attnVecPost}}

\newcommand{\prodVecCompensated}{\hat{\bm{p}}}

\usepackage[textsize=tiny]{todonotes}

\title{Top-Theta Attention: Sparsifying Transformers by Compensated Thresholding}

\author{%
Konstantin Berestizshevsky
\\
  Computing Systems Laboratory\\
  Huawei Research, Zurich, Switzerland\\
  \texttt{konstantin.berestizshevsky@huawei.com} \\
  \And
  Renzo Andri \\
  Computing Systems Laboratory\\
  Huawei Research, Zurich, Switzerland\\
  \texttt{renzo.andri@huawei.com} \\
  \AND
  Lukas Cavigelli \\
  Computing Systems Laboratory\\
  Huawei Research, Zurich, Switzerland\\
  \texttt{lukas.cavigelli@huawei.com} \\
}

\begin{document}
\makeatletter
\renewcommand{\@noticestring}{Extended version of a paper accepted at ICANN 2026.}
\makeatother

\maketitle

\begin{abstract}

We present Top-Theta (\topth) Attention, a training-free method for sparsifying transformer attention during inference. Our key insight is that static, per-head thresholds can be calibrated to retain the desired constant number of significant elements per attention row. This approach enables content-based sparsity without retraining, and it remains robust across data domains. We further introduce compensation techniques to preserve accuracy under aggressive sparsification, establishing attention thresholding as a practical and principled alternative to top-k attention. We provide extensive evaluation on natural language processing tasks, showing that \topth achieves 3–10\texttimes\ reduction in $\bm{V}$-cache usage and up to 10\texttimes\ fewer attention elements during inference while degrading no more than 1\% in accuracy.
\end{abstract}

%
\section{Introduction}\label{sec:introduction}
%
The transformer architecture has revolutionized natural language processing~\cite{vaswani2017attention} and computer vision~\cite{dosovitskiy2020image} by enabling models to capture complex dependencies through self-attention mechanisms effectively. However, despite its advantages, the attention mechanism suffers from quadratic time and linear memory complexity~\cite{keles2023computational}. Furthermore, the commonly used key-value (KV) cache optimization increases the memory requirement linearly with sequence length during the generative decoding phase. As a result, cache size requirements often exceed the physical limits of available memory~\cite{ge2024modeltellsdiscardadaptive}, and memory bandwidth becomes a major bottleneck.
Attention approximations~\cite{wang2024limitssurveytechniquesextend,fuad2023survey}, such as sparsification, promise a solution to these challenges by reducing the number of computations and the amount of data moved, focusing only on the most relevant tokens in a sequence. 
Research on sparsification has been predominantly focused on either fixed-sparsity patterns, which assume that specific token locations in the sequence are always more important (e.g., the first and last tokens in the sequence), or the content-based sparsity patterns, which require evaluating the attention scores to decide which tokens are more important (e.g., \topk attention by~\citet{gupta2021memory}).
Our work focuses on the more challenging, content-based sparsity patterns. To exploit the sparsity potential, we investigate the sparsification of the attention elements through pruning by \textit{comparing to a threshold value}. We calibrate the thresholds to select a desired average number of $k$ important tokens in every attention row. We find that calibrating model-specific thresholds is sufficient to replace the top-k search over the attention elements. Once the important tokens have been quickly determined by thresholding, the remaining tokens can be excluded from participating in the softmax computation and in the multiplication by the $\bm{V}$-matrix, thus avoiding the need to load the corresponding $\bm{V}$-rows. Moreover, to preserve the high accuracy in the downstream task, we propose numerical compensation methods such as softmax denominator compensation and mean $\bm{V}$-row compensation. 
\paragraph{Contributions} This work is taming the potential of content-based sparsity for more compute- and memory-efficient attention with negligible accuracy degradation. Our fundamental finding is:
\textit{Static thresholds can be calibrated for a given attention head and used to sparsify its attention matrices to approximately $k$ elements per row.} 
From this fundamental finding, we derive the \topth attention method, unlocking the following advantages:
\begin{enumerate}
    \item \textit{Efficiency}. 3\texttimes\ to 10\texttimes\ fewer $\bm{V}$-rows needed and 10\texttimes\  less attention elements needed for LLaMA2 and LLaMA3 models to achieve same accuracy.
    \item \textit{Tiling compatible}. Thresholding is a simple elementwise operation with no row dependency, making it applicable to tiled attention matrices required for high-performance kernels and distributed inference. 
    \item \textit{No retraining}. Threshold calibration requires only a few hundred samples and no retraining.
    \item \textit{Distribution shift resilience}. Thresholds remain consistent across input domains, representing a fundamental model characteristic that requires only one-time calibration.
\end{enumerate}
%
\section{Background}\label{sec:background}
%
%
\subsection{Transformer Models}\label{sec:background:transformer}
%
Modern Large Language Models (LLM) used for text generation primarily employ decoder-only transformer layers, noted for strong zero-shot generalization~\cite{wang2022languagemodelarchitecturepretraining} and widespread success in chatbots and productivity tools. These models operate in two phases: processing the entire input at once (prefill) and generating tokens sequentially (generative decoding). Our research aims to enhance the underlying self-attention mechanism of decoder-only transformers.
%
\subsection{Self-Attention and Sparsity}\label{sec:background:attention}
%
Multi-head self-attention (MHA) is the first computational step of the transformer layer. The MHA receives as input a sequence of tokens $\bm{X}\in\mathbb{R}^{\seqlen\times\embdim}$ where $\seqlen$ is the sequence length and $\embdim$ is the hidden dimension. Each of the heads processes $\bm{X}$ in parallel by first multiplying it by 3 different trained matrices $\bm{W_Q},\bm{W_K},\bm{W_V}\in\mathbb{R}^{\embdim\times\headdim}$ obtaining 3 matrices $\bm{Q},\bm{K},\bm{V}\in\mathbb{R}^{\seqlen\times\headdim}$ and adding a positional encoding (e.g., RoPE~\cite{su2024roformer}) to them.
 Then, a \textit{pre-softmax attention matrix} $\attnMatPre$ is computed from matrices $\bm{Q}$ and $\bm{K}$ (\ref{eqn:attn_pre_softmax}). Although $\attnMatPre$ matrix is often normalized by $\sqrt{\headdim}$ for numerical stability and masked by a causality mask, we omit these 2 steps for simplicity.
\begin{equation}\label{eqn:attn_pre_softmax}
    \attnMatPre = \bm{QK}^T
\end{equation}
 After that, each row of the $\attnMatPre$ matrix is normalized by the Softmax operation, yielding the \textit{post-softmax attention matrix} $\attnMatPost$ (\ref{eqn:attn_post_softmax}), which is then multiplied by $\bm{V}$ (\ref{eqn:sv}).
\begin{align}\label{eqn:attn_post_softmax}
    \attnMatPost&=\mathrm{row\_softmax}(\attnMatPre)\triangleq \left[ \frac{e^{A_{ij}}}{\sum_{k=1}^n e^{A_{ik}}} \right]_{\substack{1 \leq i \leq n \\ 1 \leq j \leq n}}\\
\label{eqn:sv}
    \bm{P}&=\attnMatPost\bm{V}
\end{align}
\paragraph{Prefill phase} Both pre-, and post-softmax matrices are of the shape $\seqlen\times\seqlen$, in which the element $(i,j)$ signifies the importance of token $j$ for token $i$. The pre-softmax attention $\attnMatPre$ has a range of all real numbers distributed normally, whereas the post-softmax $\attnMatPost$ has the range of $[0,1]$ and its distribution resembles log-normal, with the majority of the values concentrated near 0.
The attention elements in the initial layers exhibit a less skewed distribution (i.e., high entropy), whereas the following layers have a more concentrated distribution (i.e., low entropy) with a few rare high-attention values~\cite{vig2019analyzingstructureattentiontransformer,nahshan2024linearlognormalattentionunbiased}. 
In both $\attnMatPre$ and $\attnMatPost$, a lower attention value has a lower contribution to the further computation because $\attnMatPost$ is obtained through an order-preserving transformation of $\attnMatPre$, after which $\attnMatPost$ is multiplied by the matrix $\bm{V}\in\mathbb{R}^{\seqlen\times\headdim}$ resulting in the output matrix $\bm{P}\in\mathbb{R}^{\seqlen\times\headdim}$ (\ref{eqn:sv}). Therefore, small values in column $i$ in $\attnMatPost$ diminish the impact of row $i$ of the $\bm{V}$ matrix. \textit{This fundamental property of the attention elements allows ranking them according to their significance and pruning the least significant ones for sparsification}. 
%

%
\paragraph{Generative decoding phase and KV-cache} MHA employs a well-established performance optimization called KV-cache~\cite{shi2024costdownreviewmethods} which allows processing a single embedded token $\bm{x}\in\mathbb{R}^\embdim$, computing only the current token's $\bm{q},\bm{k},\bm{v}\in\mathbb{R}^\headdim$ vectors, while the complete $\bm{K}, \bm{V}$ matrices are loaded from the cache (avoiding recomputation) and the new $\bm{k},\bm{v}$ vectors are appended to them. In this situation, the attention matrices simplify to vectors $\attnVecPre, \attnVecPost\in\mathbb{R}^\seqlen$ representing the attention of the currently generated token to all the previous tokens, and the $\bm{P}$ matrix becomes a single token embedding $\bm{p}=\bm{s}\bm{V}\in\mathbb{R}^\headdim$. Due to the large size of the KV-caches, especially as the sequence length grows longer, the computation of the self-attention during the generative decoding is heavily limited by memory bandwidth, dominated by loading all the $\seqlen$ rows of the $\bm{K}$ and $\bm{V}$ matrices, while only performing 1 multiply-add per value read from memory. \textit{Due to the memory bottleneck, a reduction of memory reads directly leads to a corresponding speed up.}
A variant of multi-head self-attention is called grouped query attention (GQA). The GQA introduces sharing a single pair of $\bm{K},\bm{V}$ matrices for a group of $g$ heads (queries), which reduces the amount of $\bm{K},\bm{V}$ data to load by a factor of $g$. Studies have shown that GQA has a marginal impact on the downstream task accuracy, making it a favorable optimization~\cite{ainslie2023gqa}.
\subsection{\topk Attention}\label{sec:introduction:topk}
One widely adopted approach to sparsifying the attention row is finding its top $k$ out of $\seqlen$ elements and discarding the rest, where $k$ is a hyperparameter~\cite{gupta2021memory}. However, since the attention rows are often partitioned (tiling across the sequence dimension) and processed in parallel, computing the exact top-k values in a vector imposes an undesired full-row dependency, thereby constraining the tiling strategies. Moreover, the computation of the top elements requires several computational steps~\cite{zhang2023paralleltopk}, leading to a logarithmic complexity at the very best case. \textit{We conjecture that the top-k search algorithms can be replaced by comparison to a carefully calibrated threshold.}
%

%
\section{\topth Method}\label{sec:method}
%
\begin{figure}[ht]
     \centering
    \includegraphics[width=1.0\textwidth]{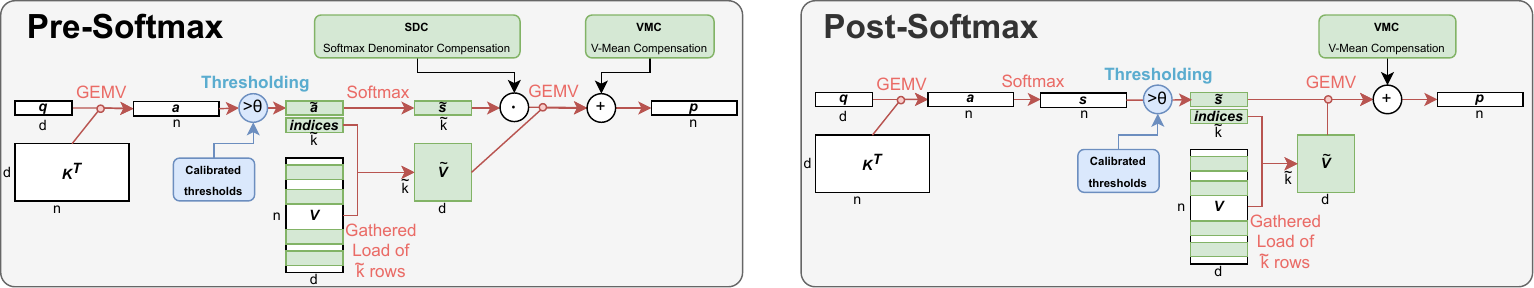}
    \caption{Two variants of \topth attention for inference at generative decoding.}
    \label{fig:top_th_inference}
\end{figure}
Our proposed \topth attention involves comparing each attention vector against a calibrated threshold. Attention elements that fall below the threshold are pruned away from subsequent computations, enhancing the efficiency and the focus of the model. Our underlying assumption is that a particular distribution of values characterizes each row of the attention matrix; therefore, using a static threshold should keep approximately the desired number of selected attention elements. Our motivation for using a threshold-based method instead of computing the top-k attention values is that thresholding is a simple elementwise operation, requiring only a constant time and involving no row dependency. In contrast, ranking the elements of a vector requires at least logarithmic time and depends on the full length of the vector. \topth can be seen as an approximate \topk, as it allows calibrating the thresholds for specific user-defined $k$.

%
\newpage
\subsection{Threshold Calibration}\label{sec:method:calibration}
%
\vspace{-5pt}
\begin{wrapfigure}{R}{0.51\textwidth}
    \vspace{-10pt}
    \begin{minipage}{0.53\textwidth}
    \begin{algorithm}[H]
    \caption{Calibrate($\calibrationSet,k,\alpha$) - 1 head threshold}
    \label{alg:calibrate}
    \begin{algorithmic}[1]
        \REQUIRE \( \calibrationSet \) (Calibration set of inputs)
        \REQUIRE \( k\in\mathbb{N} \) (elements to keep per attention row)
        \REQUIRE \( \alpha\in\mathbb{R} \) (calibration offset in std\_devs)
        \STATE $\Theta_{r}=\emptyset, \forall r $ \COMMENT{empty sets of observed thresholds}
        \FOR{\( \bm{X} \in \calibrationSet \)}
            \IF{$\mathrm{is\_prefill}(\bm{X})$} 
                \STATE $\attnMatPre = \bm{Q}(\bm{X})\bm{K}^T(\bm{X})$
                \STATE $\seqlen = \mathrm{num\_rows}(\attnMatPre)$
                \FOR{\( r = k \) to \( \seqlen - 1 \)}
                    \STATE $\Theta_{r}$ = $\Theta_{r}\cup\{\mathrm{quantile}_{\frac{\seqlen-k}{n}}(\attnMatPre_r)\}$
                    \STATE $\attnMatPre_r=\mathrm{top}_k(\attnMatPre_r)$ 
                \ENDFOR
                \STATE $\attnMatPost=\mathrm{row\_softmax}(\attnMatPre)$                
            \ELSE[generative decoding, $X\in\mathbb{R}^\headdim$]
                \STATE $\attnVecPre = \bm{Q}(\bm{X})\bm{K}^T(\bm{X})$
                \STATE $\seqlen = \mathrm{length}(\attnVecPre)$            
                \STATE $\Theta_{\seqlen}$ = $\Theta_{\seqlen}\cup\{\mathrm{quantile}_{\frac{\seqlen-k}{\seqlen}}(\bm{a})\}$
                \STATE $\attnVecPre=\mathrm{top}_k(\attnVecPre)$
                \STATE $\attnVecPost=\mathrm{softmax}(\attnVecPre)$                
            \ENDIF
        \ENDFOR
        \STATE \textbf{return} $\theta_{r}=\mathrm{mean}(\Theta_{r})+\alpha\cdot \mathrm{std\_dev}(\Theta_{r}), \forall r$
    \end{algorithmic}
    \end{algorithm}
    \end{minipage}
    \vspace{-20pt}
\end{wrapfigure}
\begin{wrapfigure}{r}{0.46\textwidth}
    \centering
    \vspace{-10pt}
    \includegraphics[width=0.46\textwidth]{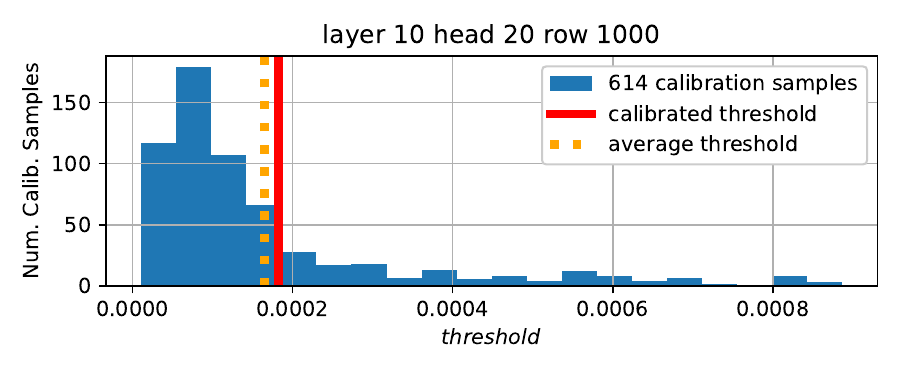}
    \caption{Distribution of the set $\Theta_{1000}$, during calibration of threshold $\theta_{10,20,800}, \alpha=0.1$ on \hell, LLaMA2-7b.}
    \label{fig:th_calibration_histograms}
    \vspace{-5pt}
\end{wrapfigure}

Prior to using \topth for efficient inference, its thresholds need to be calibrated with respect to a user-defined parameter $k$. To this end, we present~\cref{alg:calibrate} that calibrates a threshold $\theta_{r}(k)\in\mathbb{R}$ for attention row id $r$ of a given transformer layer and head. The calibration is performed offline, before the model deployment, by collecting the $(\frac{r-k}{k})$-quantile of each attention row from a set of calibration samples (l. 7), then averaging these per-sample thresholds to obtain $\theta_{r}(k)$ (l.19). This algorithm has to be performed for all layers and heads in parallel, and can be integrated into a single forward pass of the model (then disabled after the calibration is complete). We found that using a single $k$ per layer suffices for robust performance, though different $k$ can be chosen per head or row.
For practical calibration, we recommend a calibration set of a few hundred samples, as we found that increasing the calibration set size mainly improves the fidelity of thresholding to keep the desired $k$ elements per row, but does not benefit the accuracy of the downstream task (see~\cref{appendix:calib_set_size}). An optional offset hyperparameter $\alpha$ allows conservative adjustment of the final threshold. See~\cref{fig:th_calibration_histograms} for the visualization of the threshold values collected for a specific layer, head, row, and selected threshold.
%

%
%

%
During calibration, a “Top-k at calibration” step (l. 8,15) ensures that subsequent layers' activations reflect the sparsification pattern, improving threshold stability at test time. We note that \cref{alg:calibrate} in its presented form is calibrating pre-softmax thresholds, whereas it can be adapted to post-softmax by taking the quantile on the post-softmax attention $\attnMatPost,\attnVecPost$ rather than on the $\attnMatPre,\attnVecPre$. We also note that calibrating for rows $r<k$ is unnecessary for the first $k$ rows due to causal masking. 
Once the calibration is complete, the calibrated thresholds can be stored along with the model parameters, as they take negligible memory. For example, when 1200 per-row float16 thresholds are calibrated (as in \arcc) for every head, the total memory for thresholds for an entire LLaMA-3-70B model (80 layers, 64 heads) is only $80 \cdot 64 \cdot 1200 \cdot 2 = 11.8$ Mbytes (0.0008\% of the model size).

\begin{wrapfigure}{r}{0.51\textwidth}
    \centering
    \vspace{-10pt}
    \includegraphics[width=0.51\textwidth]{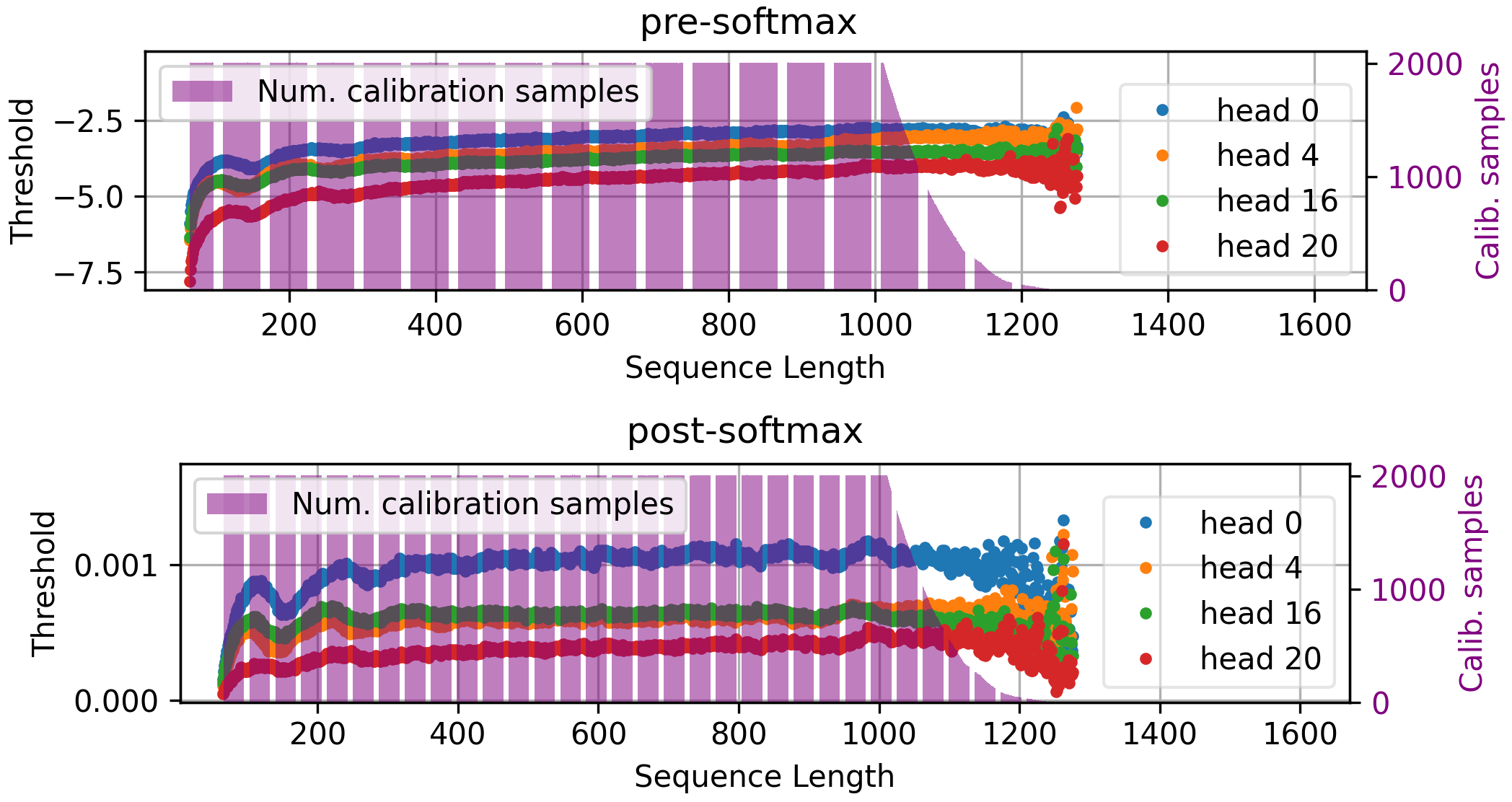}
    \caption{\textbf{Threshold values as a function of a sequence length} LLaMA2-7b,
    11\textsuperscript{th} transformer layer, calibration targeted $k=64$. 
    }
    \label{fig:th_llama}
    \vspace{-10pt}
\end{wrapfigure}
%

%
\cref{fig:th_llama} visualizes two threshold variants of the 11\textsuperscript{th} layer of LLaMA2-7b, which were calibrated for $k=64$: pre-softmax, and post-softmax.  Notably, the threshold values depend not only on the sequence length (x-axis) but also on the attention head (different colors), justifying the calibration of an individual threshold for each attention head. Secondly, thresholds obtained pre-softmax approach a constant value as the sequence length increases, whereas thresholds obtained from the post-softmax tend to decrease with longer sequence lengths. The latter is an effect of the softmax normalization that reduces the share of essential tokens from the total row sum of 1 as the sequence length increases; therefore, keeping them requires lowering the threshold. The \textit{scattering} artifact of the thresholds in the longer calibrated sequence lengths is due to the scarcity of the calibration samples in the dataset that corresponded to these sequence lengths (low purple bars in~\cref{fig:th_llama}). ~\cref{appendix:th_llama} presents a visualization of thresholds from additional layers.
\paragraph{Threshold function fitting} Our main proposal is to deploy the calibrated thresholds as complementary model parameters (one threshold parameter per every layer, head, and sequence length). However, as can be observed in~\cref{fig:th_llama}, the thresholds have a rather regular behavior as a function of the sequence length (i.e., of the attention row id). Therefore, we can fit a function to it and thereby parameterize hundreds of thresholds by a few parameters. Such a fitting can be obtained via a weighted least squares solution, with the weights being the number of calibration samples obtained for each sequence length. With this weighing, the scattering artifact observed in the longer sequence lengths will be mitigated.
\paragraph{Multi-k calibration} It is possible to calibrate multiple threshold sets $\theta_l(k_1),\theta_l(k_2),...$ targeted at different number of top elements $k_1,k_2,...$ to be preserved. Following such a calibration, one can dynamically choose the desired accuracy-performance tradeoff at inference by applying a set of thresholds with a different $k$, a desired capability for a flexible LLM serving~\cite{miao2023efficientgenerativelargelanguage}. Instead of applying ~\cref{alg:calibrate} several times with the different $k$ values, we have developed a generalized calibration algorithm, which performs a single pass over the calibration samples. The generalized procedure collects up to $n$ thresholds for every attention vector of length $n$ in the calibration sample, merges this threshold information, and eventually outputs for each (layer, head, row) a function $\theta(k)$ from $k$ to a threshold that satisfies it. Although such a calibration procedure poses large memory requirements, reducing its intensity by sampling fewer than $n$ thresholds per vector is possible (see~\cref{appendix:multik_cumulitive_calibration}).
%
\subsection{\topth Attention Inference}\label{sec:method:inference}
%
The \topth attention inference operation in~\cref{fig:top_th_inference} focuses on a single transformer layer $l$, single head $h$, and single attention row  $n$ of length $n$ (as in generative decoding). The attention vector is compared against the calibrated threshold value $\theta_{n}$. Attention elements that do not pass the threshold are discarded. Instead of multiplying by the entire $\bm{V}$ matrix, only the selected $\Tilde{k}$ row indices are loaded to a compact matrix $\Tilde{\bm{V}}$, which is used to compute the final product $\bm{p}$.
Technical details: 
(i) Since the calibration set might not have covered the entire range of sequence lengths $r\in\{k, k+1,\ldots\}$, at inference, the threshold of the nearest calibrated sequence length is used.
(ii) Threshold calibrated for a target $k$ is guaranteed to select $k$ attention elements per row only on average, i.e., actual counts ($\Tilde{k}$) may slightly vary due to input-dependent attention distributions. To handle cases where $\Tilde{k} > k$, we experimented with capping the selection at $k$ elements, prioritizing first and last tokens, but found that this noticeably degraded performance on generative tasks. Instead, we mitigated the variability of $\Tilde{k}$ by increasing the calibration set size (\cref{appendix:calib_set_size}).
The rest of this section introduces two efficient compensation mechanisms that mitigate accuracy degradation in \topth attention through improved mathematical approximation of full self-attention. 
%
\subsubsection{Softmax Denominator 
 Compensation (SDC)}
%
Since the post-softmax sparsification showed higher accuracies in our experiments compared to pre-softmax, we are interested in approximating post-softmax-sparsified attention ($\attnVecSparsePost$) using pre-softmax-sparsified attention ($\attnVecSparsePre$).
Let $\selids\subseteq\left\{0,\dots,n-1\right\}$ denote a set of indices that we intend to keep during the sparsification, and let us establish the relation between post-softmax sparsified vector $\attnVecSparsePost$ and the pre-softmax sparsified vector that underwent softmax ($\mathrm{softmax}(\attnVecSparsePre)$). Let $R$ and $E$ denote the sums of exponents of the selected and discarded elements, respectively.
\begin{align}
\forall i\in\selids, \attnVecSparsePost_i
 &= \mathrm{softmax}(\bm{a})_i = \frac{e^{a_i - \max_l{a_l}}}{\sum_{j=0}^{n-1}e^{a_j - \max_l{a_l}}} = \frac{e^{a_i - \max_l{a_l}}}{\sum_{j\in\selids}e^{a_j - \max_l{a_l}} + \sum_{j\notin\selids}e^{a_j - \max_l{a_l}}} \nonumber \\
 &= \frac{e^{a_i - \max_l{a_l}}}{R+E} = \frac{e^{a_i - \max_l{a_l}}}{R}\cdot\frac{R}{R+E} = \mathrm{softmax}(\attnVecSparsePre)_i\cdot\frac{R}{R+E}\label{eqn:sdc}
\end{align}
In other words, to achieve a post-softmax sparsification effect, one can perform pre-softmax sparsification, estimate $\Tilde{E}\approx E$, and compensate by multiplying by a factor of $R/(R+\Tilde{E})$.
The multiplication step can be applied after the softmax (\ref{eqn:attn_post_softmax}) or even after the $\bm{SV}$ product (\ref{eqn:sv}), similarly to the flash-attention~\cite{dao2022flashattention}.
We consider 3 estimations:
\begin{enumerate}
    \item \texttt{offline-calibrated}: calibrate a static compensation value $\Tilde{E}$ for every (layer, head, row) similarly to the method of threshold calibration. 
    \item \texttt{exp-threshold}: $\Tilde{E}=\gamma(n-\Tilde{k})e^\theta$ where $\theta$ is the calibrated threshold for the current sequence length $n$, and $\Tilde{k}$ is the number of selected attention elements. The intuition behind this approximation is that all the not-selected attention elements are less than $\theta$. The $\gamma$ is a small constant hyperparameter that we set to $0.05$ to approximate the difference between the sum of exponentiated thresholds and actual exponentiated discarded attention elements.
    \item \texttt{exact}: compute $\Tilde{E}=E$ by summing up exponents of the non-selected row elements. 
\end{enumerate}

Assuming that $\mathrm{argmax}_l{a_l}$ is included in the selected elements $\selids$, we do not correct the maximum value subtracted from the exponents during softmax computation.
%
\subsubsection{V-Mean-Compensation (VMC)}
%
Applying \topth in the following two cases will result in attention vector not summing up to 1: (i) post-softmax, (ii) pre-softmax followed by SDC. As a result, the value given by the product of non-selected attention elements and their corresponding $\bm{V}$-rows will be missing in the final product (\ref{eqn:sv}). We denote these cases as ``eligible for VMC'' and 
we compensate for the missing value in (\ref{eqn:sv}) by adding a mean row $\bm{\mu}$ of the $\bm{V}$ matrix scaled by the sum of all the discarded attention elements $\beta$. Equation~\ref{eqn:vmc:final} lists the computation of the corresponding $\bm{V}$-mean-compensated product $\hat{\bm{p}}\approx\attnVecPost\bm{V}$, where $\bm{\Tilde{\attnVecPost}}\in[0,1]^{\Tilde{k}}$ is a post-softmax attention vector that underwent \topth (containing the selected $\tilde{k}$ out of $n$ attention elements) and is eligible for VMC, and $\bm{\tilde{V}}$ contains selected $\tilde{k}$ rows of $\bm{V}$.
%
\begin{align}\label{eqn:vmc:mu}
    \bm{\mu} &= \frac{1}{\seqlen}\sum_{i=0}^{\seqlen-1}\bm{V}_{i}\\
\label{eqn:vmc:beta}
    \beta &= 1-\sum_{i=0}^{\Tilde{k}-1} \tilde{\attnVecSparsePost}_i \\
    \label{eqn:vmc:final}
    \prodVecCompensated &= \attnVecSparsePost V + \beta\bm{\mu}
\end{align}
%
%
We formally justify the VMC approximation in~\cref{appendix:vmc}. The intuition behind it is to approximate each of the $\seqlen-\numKept$ discarded attention values by an average and to multiply it by an average row of the $\bm{V}$ matrix. Such compensation can improve as the sequence length $\seqlen$ increases because the number of averaged elements in $\bm{\mu}$ and $\beta$ will increase accordingly. 
During generative decoding, $\bm{\mu}$ can be maintained as a running mean to avoid full recalculation.


%
\section{Evaluations}\label{sec:evaluations}
%
We use the LM Evaluation Harness~\cite{gao2024lm} to evaluate the normalized accuracy metric on multiple-choice Q\&A datasets, including their standard few-shot settings. In addition, we assess the \humaneval dataset using the official evaluation harness~\cite{chen2021codex} to measure the pass@1 metric on all 164 code generation tasks, and two long sequence summarization tasks of LongBench (qmsum and gov\_report) using their official evaluation setup~\cite{bai2024longbench}. Our hardware setup requires GPUs with a total of 192GB (to run 70B models), and a CPU with 300GB of physical memory. Reproducing the 7B and 8B model results on smaller datasets (i.e., not LongBench) can be accomplished using a single 24GB VRAM GPU. For detailed software versions and reproducibility scripts, we make our source code available\footnote{\tiny{\url{https://github.com/huawei-csl/top-theta-attention}}}.

We evaluate 3 main attention variants:
\begin{inparaenum}[(i)]
    \item Baseline -- full attention, without any sparsification;
    \item \topk\ -- keep $k$ attention elements per row;
    \item \topth\ -- keep $\Tilde{k}\approx k$ attention elements by thresholding.
\end{inparaenum}
We also apply our compensation methods to the \topk baselines for a fair comparison.
%
\subsection{\topth Overall Performance}\label{sec:evaluations:overall}
%
This section compares the best configurations of \topth against the non-sparsified baseline and \topk. 
\begin{figure}[h!]
\centering
\begin{subfigure}{0.32\textwidth}
    \includegraphics[width=\textwidth]{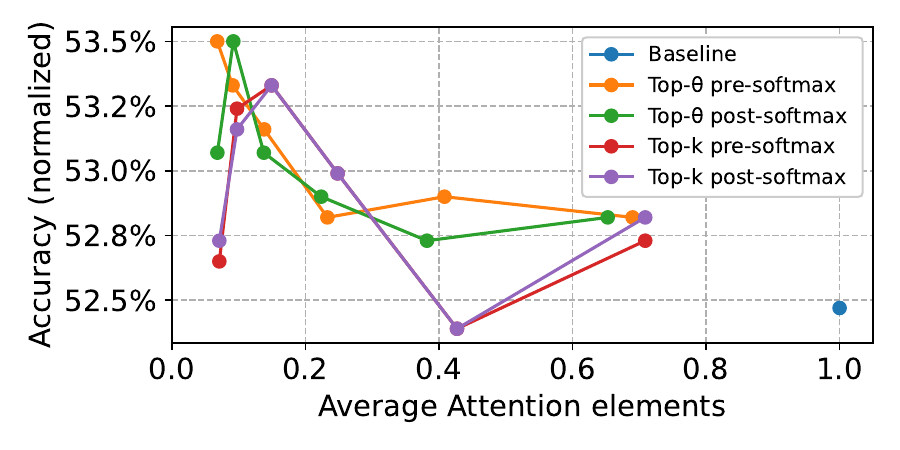}
        \caption{LLaMA2-7b \arcc}
    \label{fig:evaluations:acc_kept_attn:llama2_arc_challenge}
\end{subfigure}
\hfill
\begin{subfigure}{0.32\textwidth}
    \includegraphics[width=\textwidth]{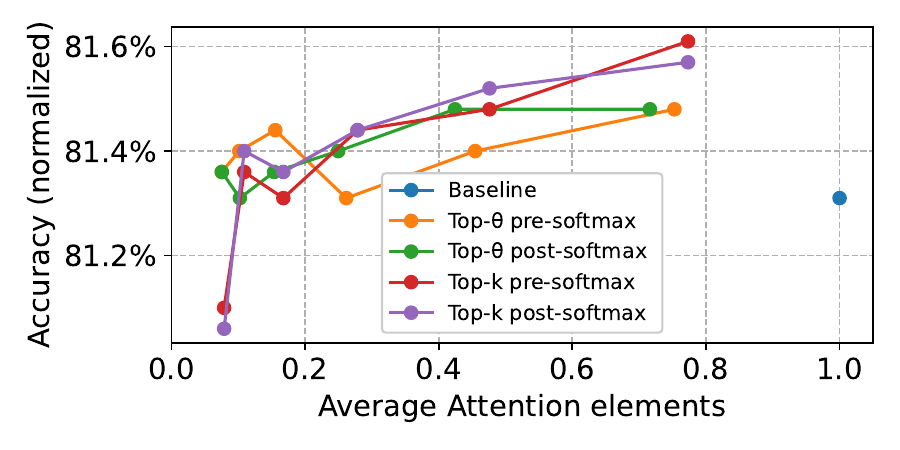}
        \caption{LLaMA2-7b \arce}
    \label{fig:evaluations:acc_kept_attn:llama2_arc_easy}
\end{subfigure}
\hfill
\begin{subfigure}{0.32\textwidth}
    \includegraphics[width=\textwidth]{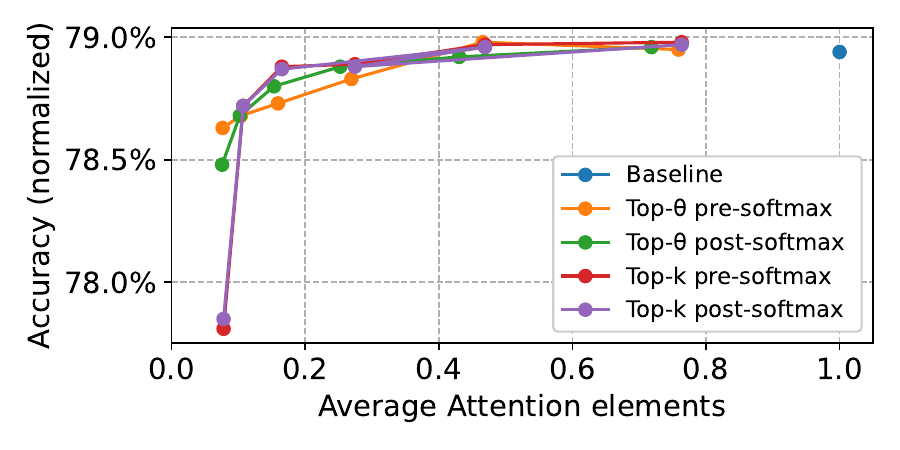}
        \caption{LLaMA2-7b \hell }
    \label{fig:evaluations:acc_kept_attn:llama2_hellaswag}
\end{subfigure}

\begin{subfigure}{0.32\textwidth}
    \includegraphics[width=\textwidth]{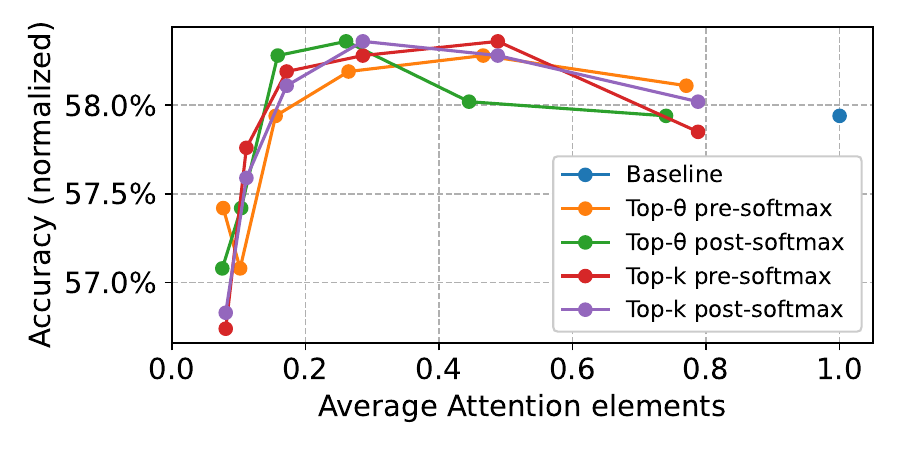}
        \caption{LLaMA3-8B \arcc}
    \label{fig:evaluations:acc_kept_attn:llama3_arc_challenge}
\end{subfigure}
\hfill
\begin{subfigure}{0.32\textwidth}
    \includegraphics[width=\textwidth]{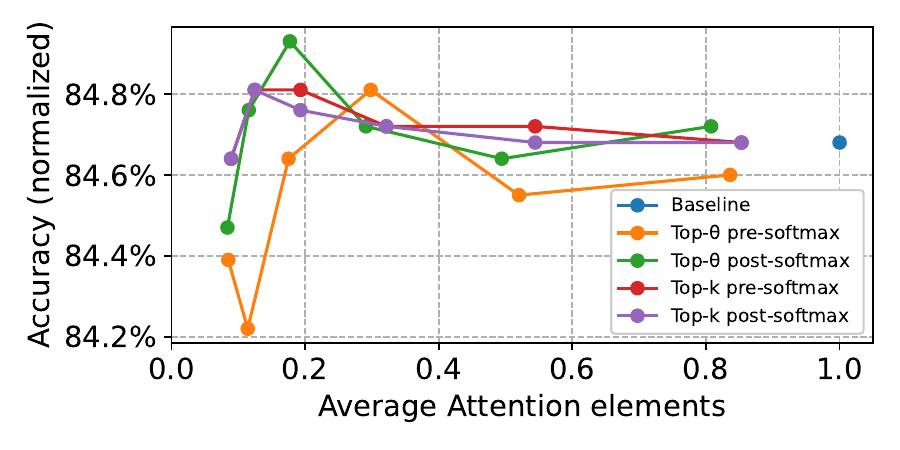}
        \caption{LLaMA3-8B \arce}
    \label{fig:evaluations:acc_kept_attn:llama3_arc_easy}
\end{subfigure}
\hfill
\begin{subfigure}{0.32\textwidth}
    \includegraphics[width=\textwidth]{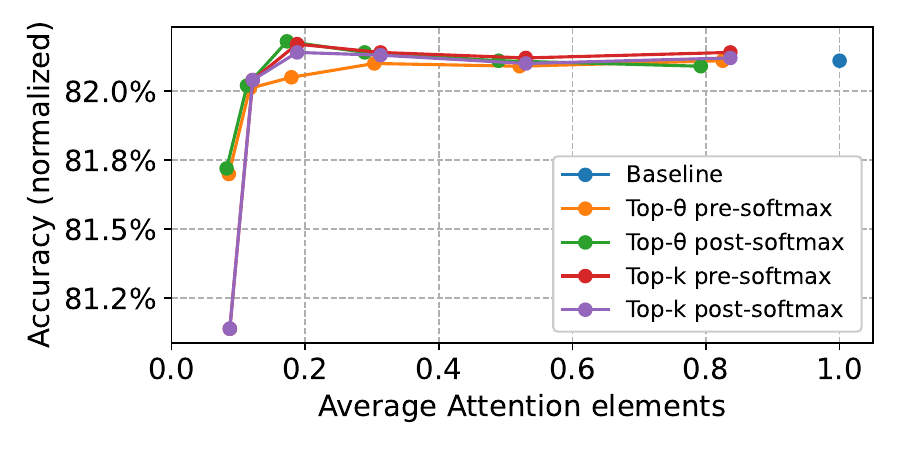}
        \caption{LLaMA3-8B \hell }
    \label{fig:evaluations:acc_kept_attn:llama3_hellaswag}
\end{subfigure}

\begin{subfigure}{0.32\textwidth}
    \includegraphics[width=\textwidth]{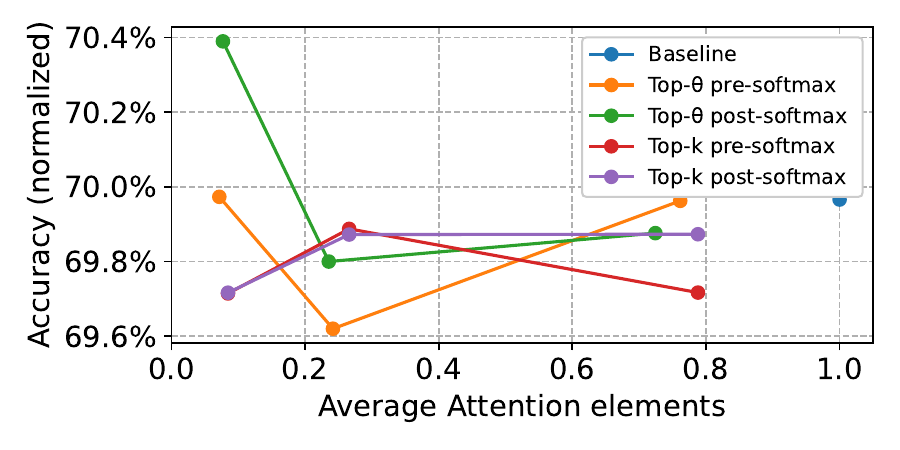}
        \caption{LLaMA3-70B \arcc}
    \label{fig:evaluations:acc_kept_attn:llama3_70_arc_challenge}
\end{subfigure}
\hfill
\begin{subfigure}{0.32\textwidth}
    \includegraphics[width=\textwidth]{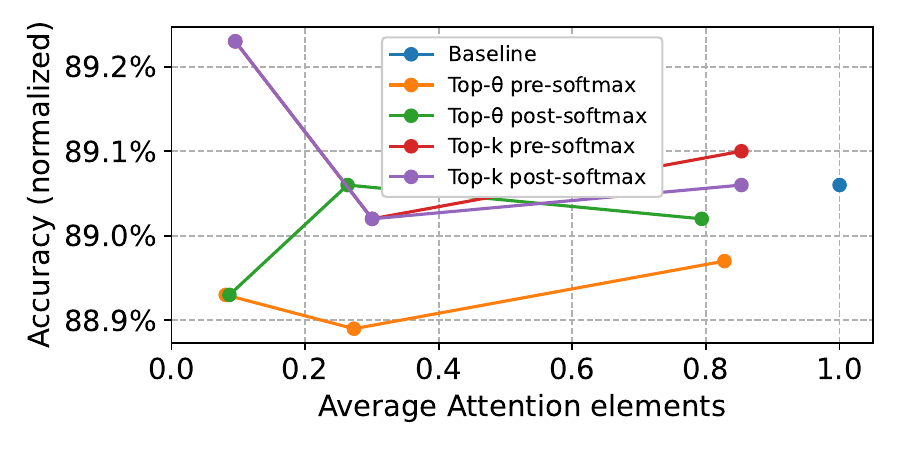}
        \caption{LLaMA3-70B \arce}
    \label{fig:evaluations:acc_kept_attn:llama3_70_arc_easy}
\end{subfigure}
\hfill
\begin{subfigure}{0.32\textwidth}
    \includegraphics[width=\textwidth]{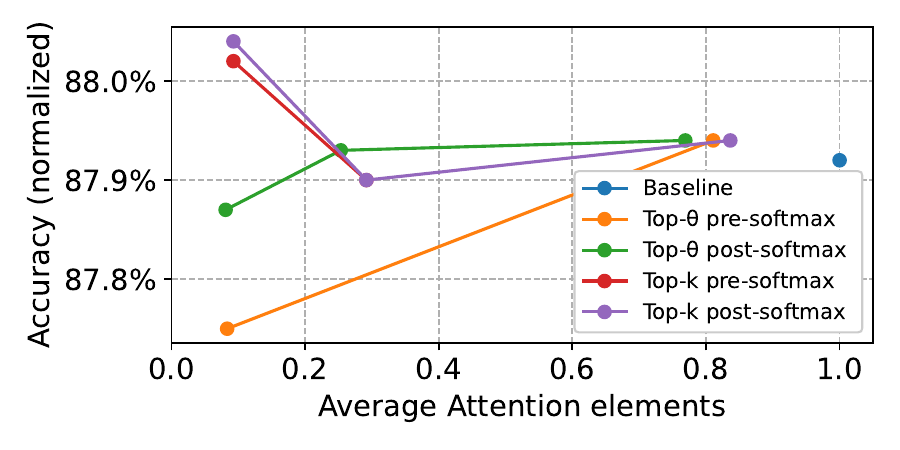}
        \caption{LLaMA3-70B \hell}
    \label{fig:evaluations:acc_kept_attn:llama3_70_hellaswag}
\end{subfigure}

\caption{\textbf{Prefill-based tasks} - Tradeoff between model accuracy (y-axis) and the portion of kept attention elements per attention head (x-axis). All post-softmax \topk and \topth employ VMC, and all pre-softmax variants employ both VMC and exact SDC. These compensations achieve little, if any, accuracy degradation while achieving up to 10\texttimes\ reduction in the attention elements. }
\label{fig:evaluations:acc_kept_attn}
\end{figure}
In~\cref{fig:evaluations:acc_kept_attn}, LLaMA2~\cite{touvron2023llama} and LLaMA3~\cite{grattafiori2024llama} models are evaluated on Q\&A tasks, where the $k$ parameter of \topk and of the \topth is swept from 32 to 512. The first two layers are kept at $k=512$. The calibration set used for \topth in each dataset is $10\%$ of the training or validation sets (different from the test set). The $x$-axis shows the fraction of the attention elements that are involved in computation, normalized to an average number of attention elements in the entire model in a given forward pass. The $y$-axis shows the normalized accuracy. The main observation is that in all models,\textit{ both \topk and \topth have increased the accuracy} (by $0.2\%-1\%$) compared to the baseline while pruning away a significant portion of attention elements ($2\times$ -- $5\times$ fewer elements were active). Secondly, \emph{post-softmax sparsification performs consistently better in both \topk and \topth, compared to the pre-softmax sparsification}.

\begin{figure}[h!]
\centering
\begin{subfigure}{0.46\textwidth}
    \includegraphics[width=\textwidth]{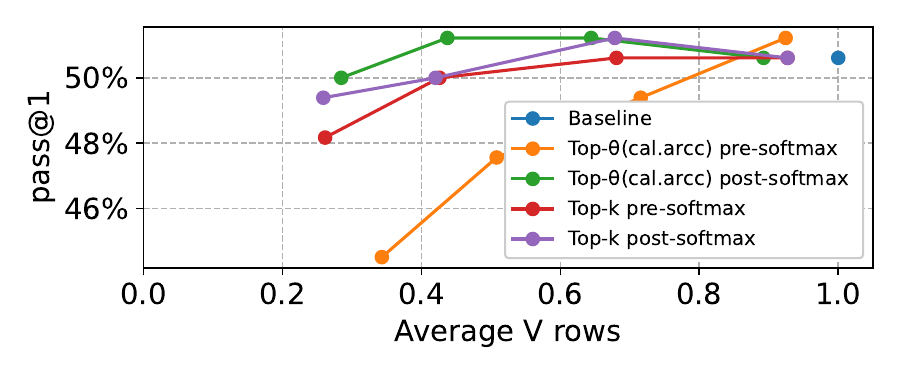}
        \caption{LLaMA3-8B-Instruct - HumanEval}
        \label{fig:evaluations_generative:acc_kept_vrow:llama3_8b_humaneval}
\end{subfigure}
\hfill
\begin{subfigure}{0.46\textwidth}
    \includegraphics[width=\textwidth]{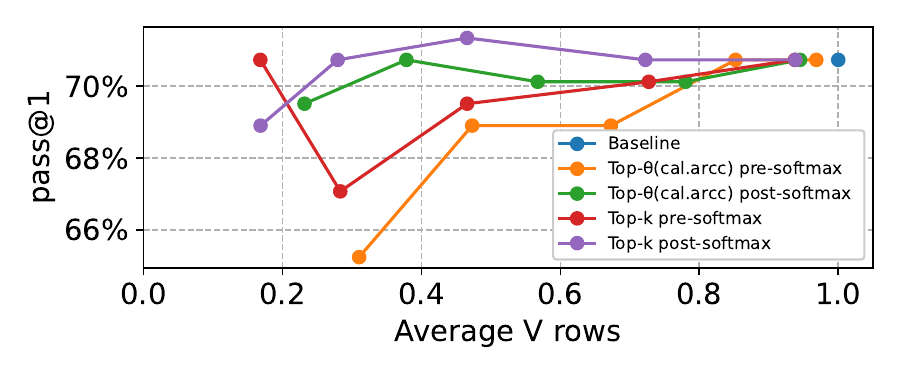}
        \caption{LLaMA3-70B-Instruct - HumanEval}
        \label{fig:evaluations_generative:acc_kept_vrow:llama3_70b_humaneval}
\end{subfigure}
\hfill
\begin{subfigure}{0.46\textwidth}
    \includegraphics[width=\textwidth]{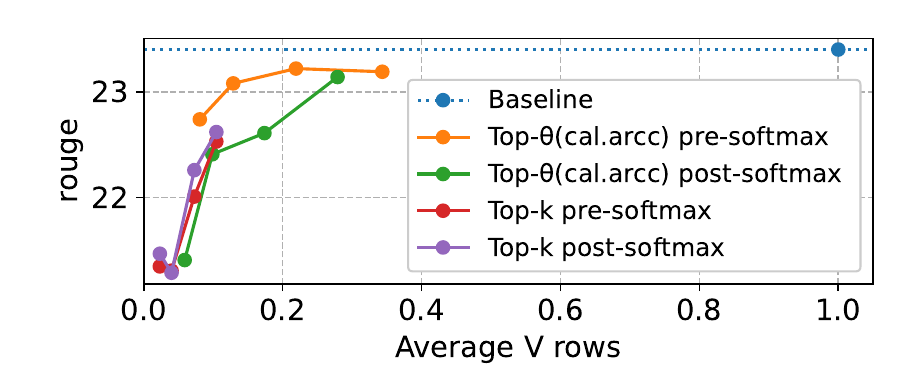}
        \caption{LLaMA3.1-8B-Instruct LongBench-qmsum}
    \label{fig:evaluations_generative:acc_kept_vrow:llama3.1_8B_qmsum}
\end{subfigure}
\hfill
\begin{subfigure}{0.46\textwidth}
    \includegraphics[width=\textwidth]{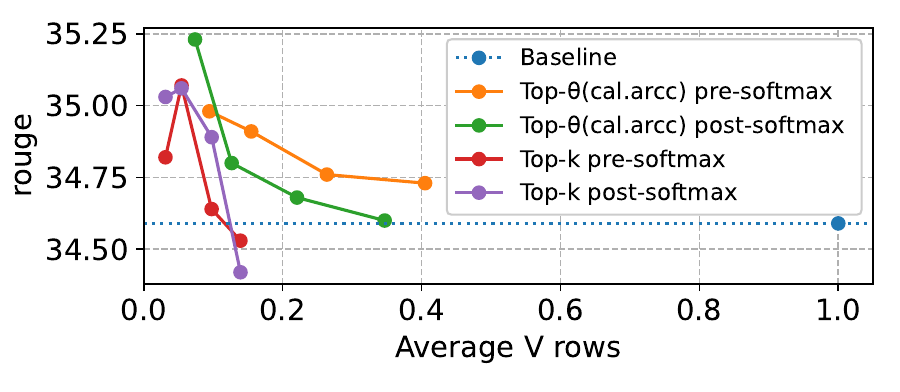}
        \caption{LLaMA3.1-8B-Instruct LongBench-gov\_report}
    \label{fig:evaluations_generative:acc_kept_vrow:llama3.1_8B_gov_report}
\end{subfigure}

\caption{\textbf{Generative Tasks} -- Tradeoff between model accuracy (y-axis) and the portion of required $\bm{V}$-rows per group of heads (x-axis). The \topth variants employ a threshold calibrated on \arcc dataset. All post-softmax \topk and \topth employ VMC, and all pre-softmax variants employ VMC and exact SDC. 3\texttimes\ and 10\texttimes\ reduction of $\bm{V}$ rows is achieved on \humaneval and LongBench, respectively.}
\label{fig:evaluations_generative:acc_kept_attn}
\end{figure}

In~\cref{fig:evaluations_generative:acc_kept_attn}, we focus on the generative tasks, where the main bottleneck is the reading of the KV cache. 
To demonstrate domain adaptation capabilities of the calibrated \topth thresholds, they were first calibrated on a different dataset (\arcc) and then loaded for \humaneval and LongBench evaluation. We examine LLaMA-3-Instruct and LLaMA-3.1-Instruct models, and we use 0 temperature for generation. 
On \humaneval task (~\cref{fig:evaluations_generative:acc_kept_vrow:llama3_8b_humaneval,fig:evaluations_generative:acc_kept_vrow:llama3_70b_humaneval}) we show the tradeoff between the output quality (pass@1) and the number of required $\bm{V}$-rows. The $k$ parameter was swept between 32 and 512 with the first two layers set at 512. The \emph{post-softmax \topth outperforms other approaches in both 8B and 70B models}, preserving pass@1 within $1\%$ of the baseline while reducing the required $\bm{V}$-rows by $3\times$. Impressively, in the 70B model, \topth offered a 5\texttimes\ reduction at the expense of $1\%$ of the pass@1. 
On LongBench tasks (\cref{fig:evaluations_generative:acc_kept_vrow:llama3.1_8B_qmsum,fig:evaluations_generative:acc_kept_vrow:llama3.1_8B_gov_report}), where the average prompt lengths reached around 15k tokens, the $k$ parameter was swept in between 128 and 768 with the first two layers set at 768, and \textit{\topth showed even higher reductions of the required $\bm{V}$-rows - up to 10\texttimes\, despite the GQA}. Such a reduction was achieved using thresholds calibrated for as few as $k=128$ elements. The Rouge-L score (higher is better) stayed within $1\%$ of the baseline, often improved by $0.5\%$.
Overall, both \topth and \topk performed similarly well on both Q\&A and generative tasks.
For a version of~\cref{fig:evaluations:acc_kept_attn,fig:evaluations_generative:acc_kept_attn} with standard deviations of the collected metrics, the reader can refer to~\cref{appendix:evaluation_statistics}. The rest of this section shows an ablation study of our method.
\subsection{Pre- vs Post-Softmax Thresholding}
\begin{wrapfigure}{r}{0.46\textwidth}
    \centering
    \includegraphics[width=0.46\textwidth]{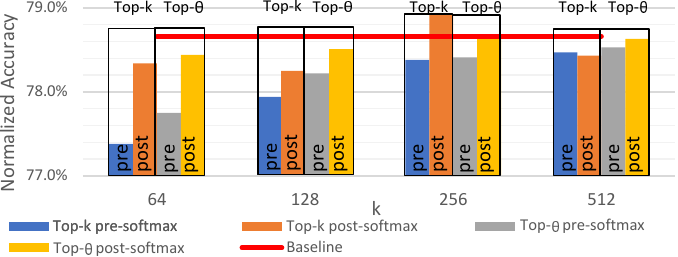}
    \caption{Pre- vs. post-softmax Top-k/$\theta$}
    \label{fig:pre-post}
    \end{wrapfigure}
We evaluate the impact of attention matrix sparsification in its two main variants: on matrix $\attnMatPre$ and on the post-softmax matrix ($\attnMatPost$), where each of these two variants requires individual calibration. \cref{fig:pre-post} depicts how the different thresholdings impact the accuracy of LLaMA2-7b on the \hell dataset. For comparison, the \topk approach is also evaluated alongside \topth. We conclude that \textit{post-softmax preserves more accuracy compared to pre-softmax thresholding}, and we provide extended results on additional datasets (see~\cref{appendix:pre-vs-post-softmax}).
%
\subsection{Thresholding Different Layers}\label{sec:evaluations:layers}
%
\vspace{-5pt}
\begin{wrapfigure}{r}{0.46\textwidth}
    \centering 
        \includegraphics[width=0.46\textwidth]{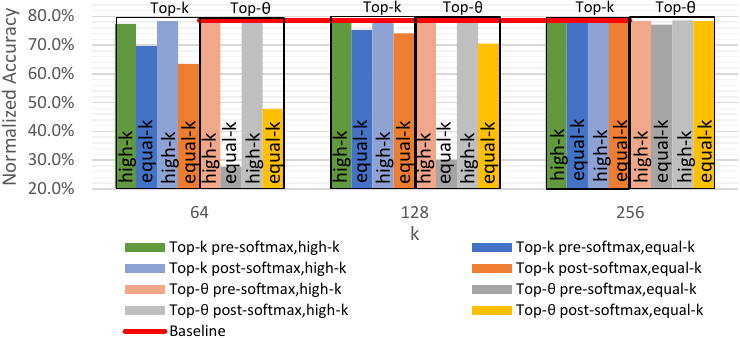} 
    \caption{High-k in first two layers}
    \label{fig:highK} 
    \end{wrapfigure}
We explored the impact of thresholding different layers with a different target $k$. For the LLaMA models, we have observed that targeting a slightly higher $k$ in the first layers is crucial. As a best practice, we found keeping 2 initial layers at $k=512$, whereas the rest of the layers could be sparsified more aggressively.~\cref{fig:highK} shows the LLaMA2-7b model accuracy on \hell, comparing \topk and \topth using higher $k$ in first 2 layers against using equal $k$ in all layers. All variants do not perform any compensations. \textit{The conclusion is that preserving denser attention matrices in the early layers is beneficial for the accuracy of a downstream task, which is aligned with some of the works on quantization}~\cite{tang2024quest,huang2024quantization}.

%
\newpage
\subsection{Thresholding Different Attention Rows}\label{sec:evaluations:rows}
\begin{wrapfigure}{r}{0.46\textwidth}
    \centering 
        \includegraphics[width=0.46\textwidth]{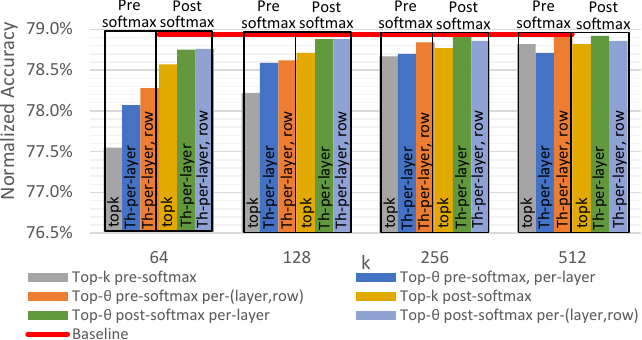} 
    \caption{Calibrating per-attention-row thresholds vs a unified threshold for all attention rows}  
    \label{fig:calibration_per_row} 
\end{wrapfigure}
In~\cref{sec:evaluations:layers}, we have shown that \topth attention should use individually calibrated thresholds for every transformer layer in the model due to inherently different attention element distributions across the layers, and in~\cref{sec:method:calibration}, we saw that different heads require individual thresholds. However, it is less obvious whether the different rows of the attention matrix require different thresholds within a single head. Evaluations of the LLaMA2-7b model on \hell (\cref{fig:calibration_per_row}) show that calibrating individual thresholds per-attention row is beneficial, especially for the pre-softmax setting. For more supporting results, refer to~\cref{appendix:evaluations:rows}.
%

%
\subsection{Numerical Compensations}\label{sec:evaluations:numerical_compensations}
%
\begin{wrapfigure}{r}{0.46\textwidth}
    \centering 
        \includegraphics[width=0.46\textwidth]{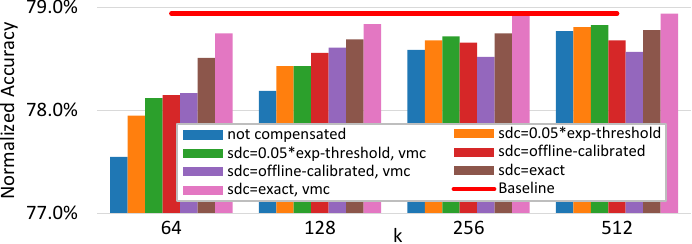} 
    \caption{\textbf{SDC and VMC compensations} impact the accuracy positively.}  
    \label{fig:numerical_compensations} 
    \end{wrapfigure}
We evaluate the proposed numerical compensation methods SDC and VMC, finding that more explicit SDC variants (exp-threshold, exact) substantially recover degraded accuracy on the challenging \hell task with LLaMA2-7b and \topth attention, as shown in~\cref{fig:numerical_compensations}; combining SDC with VMC further improves results, effectively closing the accuracy gap between the baseline and the non-compensated \topth attention. However, on the \arcc and \arce datasets, \topth attention alone already outperforms the baseline by reducing attention noise, so applying compensations in these cases offers little benefit and may even reduce accuracy back to baseline levels by reintroducing noise. \textit{Overall, SDC and VMC compensations almost entirely recover the accuracy.}
%

%
\subsection{Impact of Grouped Query Attention (GQA)}\label{sec:evaluations:impactofgqa}
%
%
\begin{wrapfigure}{r}{0.46\textwidth}
    \centering 
        \includegraphics[width=0.435\textwidth]{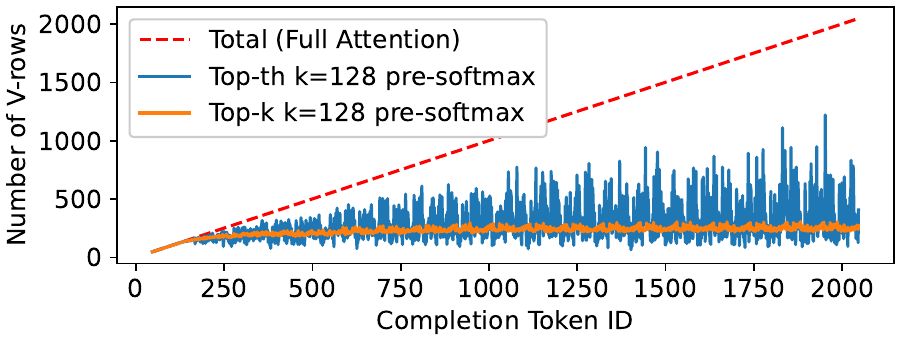} 
    \caption{\textbf{GQA impact} - Number of required $\bm{V}$-rows for every generated token, LLaMA-3-8B-Instruct (layer 2, first  GQA group), \humaneval task \#25.}  
    \label{fig:evaluations:gqa} 
    \end{wrapfigure}
%

%
In GQA, multiple attention heads share the same $\bm{V}$ matrix; for LLaMA-3-8B models with a group size of $g=4$, this means up to $4k$ $\bm{V}$ rows could be needed if heads select completely different tokens. However, as shown in~\cref{fig:evaluations:gqa}, both \topk and \topth with $k=128$ typically select only 250–300 $\bm{V}$ rows per group, indicating substantial agreement among heads - a pattern observed across many layers and further illustrated in~\cref{appendix:evaluations:impactofgqa}.
In our experiments, each head in the group used only its own selected attention elements and $\bm{V}$-rows, but two alternative approaches could improve GQA sparsification: (1) discarding $\bm{V}$-rows requested by only a few heads, and (2) augmenting each head's selections with the group's union to leverage $\bm{V}$-rows already loaded for the group and potentially improve accuracy.
%
\subsection{Distribution Shifts}\label{sec:evaluations:distribution_shift}
\begin{figure}[h!]
\centering
\begin{subfigure}{0.46\textwidth}
    \includegraphics[width=\textwidth]{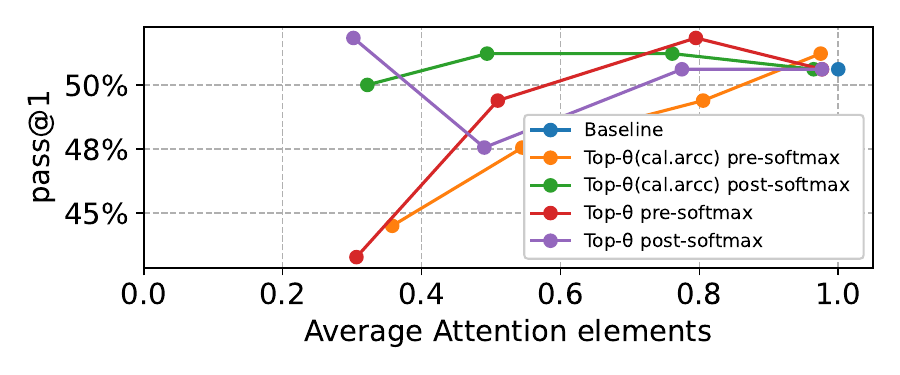} 
    \caption{LLaMA3-8B-Instruct - \humaneval}  
        \label{fig:evaluations:distribution_shift:humaneval} 
\end{subfigure}
\hfill
\begin{subfigure}{0.46\textwidth}
    \includegraphics[width=\textwidth]{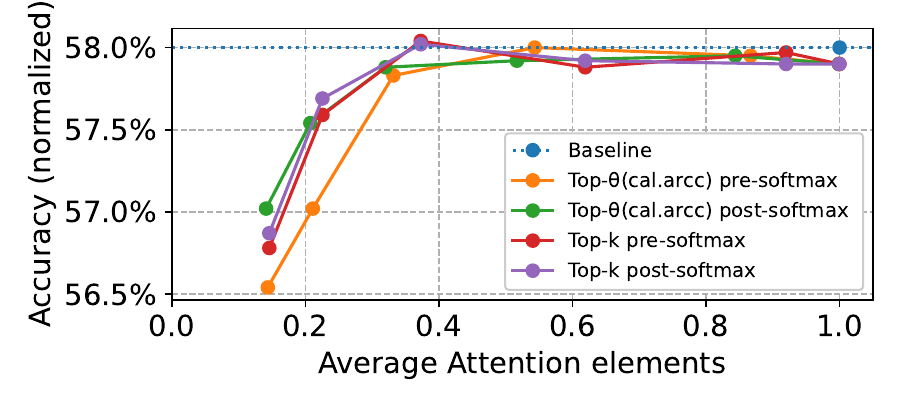}
    \caption{LLaMA3-8B MedMCQA}
        \label{fig:evaluations:distribution_shift:medmcqa}
\end{subfigure}

\caption{\textbf{Distribution shift} - \topth calibrated on different task (labeled with \textit{cal.arcc}) shows comparable accuracy and attention reduction compared to \topth calibrated on the same task~\ref{fig:evaluations:distribution_shift:humaneval}, and compared to \topk~\ref{fig:evaluations:distribution_shift:medmcqa}.}
\label{fig:evaluations:distribution_shift}
\end{figure}

%
To examine how domain-sensitive the calibrated threshold is, we evaluate on \humaneval the two following \topth variants: 1) calibrated on Q\&A dataset of \arcc, and  2) calibrated on the first 10\% of \humaneval tasks. As seen in~\cref{fig:evaluations:distribution_shift:humaneval}, the \topth post-softmax is even more accurate when using thresholds calibrated on a different dataset. For pre-softmax, there is a benefit of calibrating on the same dataset. We also evaluate using MedMCQA, where we compare the \topth calibrated on \arcc with \topk on~\cref{fig:evaluations:distribution_shift:medmcqa} - showing that both sparsification methods perform equally well. \textit{Overall, the thresholds show resilience towards distribution shift, suggesting that they are strongly associated with the model rather than with the data. This allows calibrating thresholds once per model.}
%
%
%

%
\vspace{-3mm}
\subsection{Top-Theta kernel}\label{sec:evaluations:kernel}
%
We implemented a prototype \topth MHSA kernel on an Ascend NPU (24 Davinci cores, 376\,TOPS/sec FP16, 1.8\,TB/sec memory bandwidth), achieving a $1.17\times$ speedup over the optimized \texttt{IncrementalFlashAttentionV4} decoding kernel~\cite{npu2024a} at 48 heads, $D{=}128$, $k{=}128$, and $n{=}2048$. Sparse kernel efficiency is inherently hardware-dependent, relying on low-level DMA, bandwidth, and parallel execution capabilities. To fully unlock the potential of sparsity-based attention, future memory subsystems with native gather support and compact KV-cache layouts would translate the theoretical bandwidth reductions of \topth more directly into wall-clock speedups, a trend already anticipated in the memory systems literature~\cite{miao2019streambox,chatterjee2014managing}.

%
\section{Related Work}\label{sec:related_work}
%

%
A seminal approach in \textit{content-based} sparsity is \topk attention~\cite{gupta2021memory}, which selects the $k$ largest elements per row of the attention score matrix. However, implementing efficient top-k selection is challenging in tiled kernel settings due to row-wise dependencies. Energon~\cite{zhou2022energon} addresses memory bandwidth by iteratively quantizing and filtering $\bm{K}$, but this restricts accurate computation of the final attention output, mirroring limitations seen when compensation is omitted in \topth. Other methods, such as SpAtten~\cite{wang2021spatten} and A\textsuperscript{3}~\cite{ham2020a3acceleratingattentionmechanisms}, require specialized hardware to accelerate top-k search, which \topth avoids. SparQ~\cite{ribar2023sparq} also employs top-k selection and introduces VMC-style compensation for omitted $\bm{V}$ rows, while the Swin Transformer~\cite{liu2021swin} adopts a fixed local sparsity pattern, restricting each token’s attention to its neighboring $M \times M$ patches.
Learned Token Pruning (LTP)~\cite{kim2022learned} is closely related to our work, leveraging a model-trained threshold to prune tokens in the attention matrix and achieving strong efficiency gains, but requiring full model retraining. In contrast, our approach identifies an inductive threshold within a pretrained model, enabling immediate deployment on large models without costly retraining or fine-tuning. Similarly, ~\citet{lou2024sparserfastermoreefficient} select top-k elements but also depend on retraining. Methods like Sparsemax~\cite{pmlr-v48-martins16} and Entmax~\cite{peters2019sparsesequencetosequencemodels} achieve sparsity via dynamic or iterative thresholding, but incur higher computational overhead due to sequence-wide sorting or iterative refinement. ReLA~\cite{zhang2021sparseattentionlinearunits} induces sparsity through ReLU and normalization, but lacks sparsity control and is more expensive than static thresholding.
%

Numerous works focused on \textit{fixed attention} sparsity. For example, in Longformer~\cite{beltagy2020longformer}, only a few manually picked ``global attention'' tokens are allowed to attend to other tokens, whereas other tokens are restricted to the neighboring tokens in a dilated sliding window. Our work, on the other hand, capitalizes on identifying content-based sparsity in the attention matrix, as it offers higher accuracy gains. The work of Kim and Cho~\cite{kim2021lengthadaptivetransformertrainlength} requires retraining, although it accommodates dynamic shortening of the sequence on demand at inference time.

%
\section{Limitations}\label{sec:limitations}
%
This work targets attention sparsification and reduction of value-matrix accesses with a focus on maintaining accuracy. We do not extensively assess wall-clock runtime improvements, as meaningful speedups require specialized, hardware-aware kernel optimizations, which are beyond this paper's scope. Performance depends heavily on target hardware due to varying efficiency in handling sparse memory accesses, and we strongly believe this aspect warrants a separate in-depth study. Our experiments are limited to decoder-only transformer models within the LLaMA family, which cover common attention mechanisms such as GQA and MHSA that span across some of the most successful models (Mistral, QWEN, Phi, etc.). Extending the approach to larger or different transformer architectures and other attention methods remains an important direction for future work. Additionally, factors such as model size, sequence length, and attention patterns may influence the trade-offs between sparsity and accuracy in other settings.

\section{Conclusion}\label{sec:conclusion}
%
We have presented \topth, a new sparse attention algorithm based on fixed and calibrated thresholds. At inference time, it boils down to a simple elementwise operation, which overcomes the limitation of full-row dependent algorithms, such as \topk, and thereby unconstrains tiling and distributed inference implementations. Moreover, only a small subset of calibration samples is required to calibrate the thresholds, which we showed to be resilient to distribution shifts. Therefore, a short calibration is needed once per model. Furthermore, we show minor to no performance degradation using 10\texttimes\ less attention elements at the computationally-bound prefill phase and using 3\texttimes\ (for shorter sequence tasks) to 10\texttimes\ less $\bm{V}$ matrix rows (for longer sequence tasks) at the memory bandwidth-bound generative decoding phase. These reductions unlock promising speedups for inference.

\newpage
\bibliographystyle{unsrtnat}  
\bibliography{references}


\appendix

\clearpage

%
\section{Impact Statement}\label{appendix:impact_statement}
This paper presents work whose goal is to advance the field of efficient Machine Learning. All potential societal consequences are mostly unrelated to the specific work but are more related to machine learning applications in general. One potential positive consequence of our work is that LLM technologies can be adopted by vendors and systems with lower resources, since our proposal unlocks deployment of LLMs in systems with lower memory bandwidth. We clearly and transparently state that our method is a lossy method (i.e., the accuracy of the downstream LLM task may degrade in favor of higher performance). Lossy deep learning models are a known practice in literature (e.g., when applying quantization and pruning), hence one should employ sufficient guardrails when deploying a lossy method in security-sensitive scenarios.
%

%
\section{Threshold Calibration Set Size}\label{appendix:calib_set_size}
\begin{figure}[ht]
    \centering
    \includegraphics[width=1.0\textwidth]{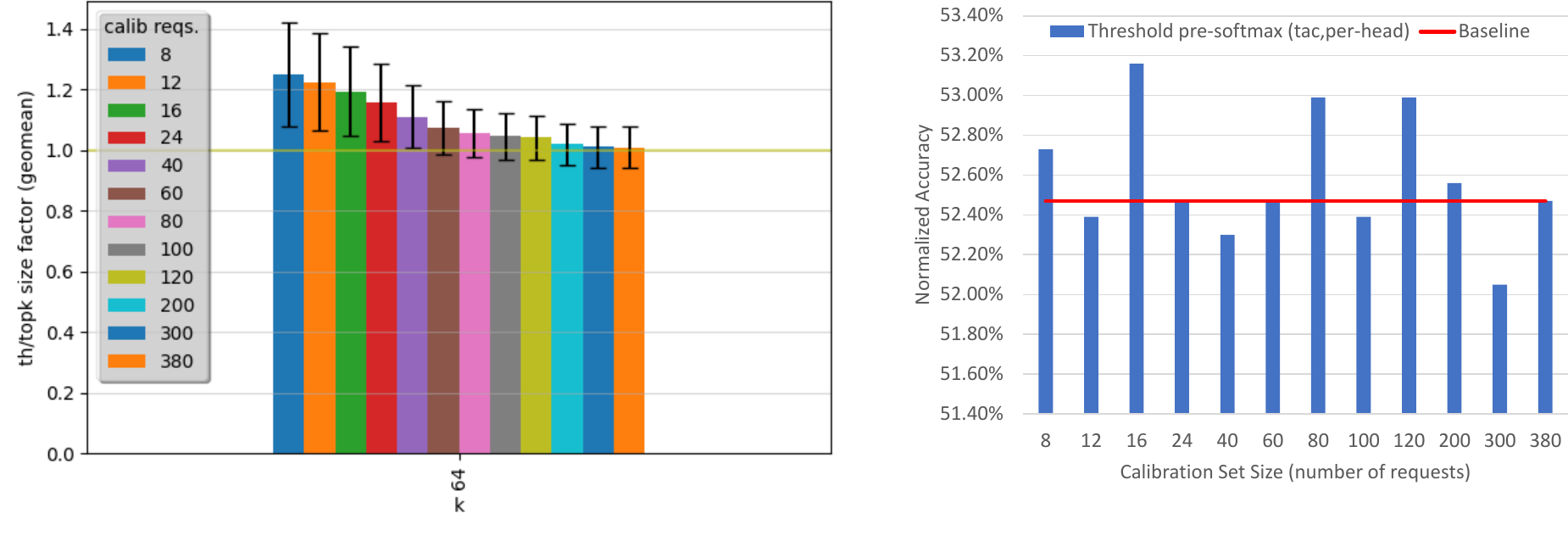}
    \caption{\textbf{Calibration set size impact} - LLaMA2-7B, calibration and evaluation on \arcc. \textbf{Left}: the Y axis shows the average ratio $\Tilde{k}/k$ (the closer to 1 the better the approximation of topk) and the X-axis (different bars) refers to different calibrations that were performed with calibration set sizes ranging from 8 to 380. \textbf{Right}: accuracy of LLaMA2-7b with \topth on \arcc dataset as a function of the calibration set size used to calibrate the thresholds.}
    \label{fig:appendix:calib_set_size}
\end{figure}

%
We have experimented with various calibration set sizes, and the main finding was that even with as few as 8 calibration samples, the model retains good accuracy. As the calibration set size grows larger, to a few hundred, the number of attention elements selected via thresholding ($\Tilde{k}$) is approaching the desired $k$ on average, with a smaller variance. See~\cref{fig:appendix:calib_set_size} where we plot on the left the ratio between the effective number of elements that passed the threshold ($\Tilde{k}$) and the desired $k$ (the closer to $1.0$ the better).

%
\section{Threshold Calibration}\label{appendix:th_llama}
\begin{figure}[h!]
\centering
\begin{subfigure}{0.46\textwidth}
    \includegraphics[width=\textwidth]{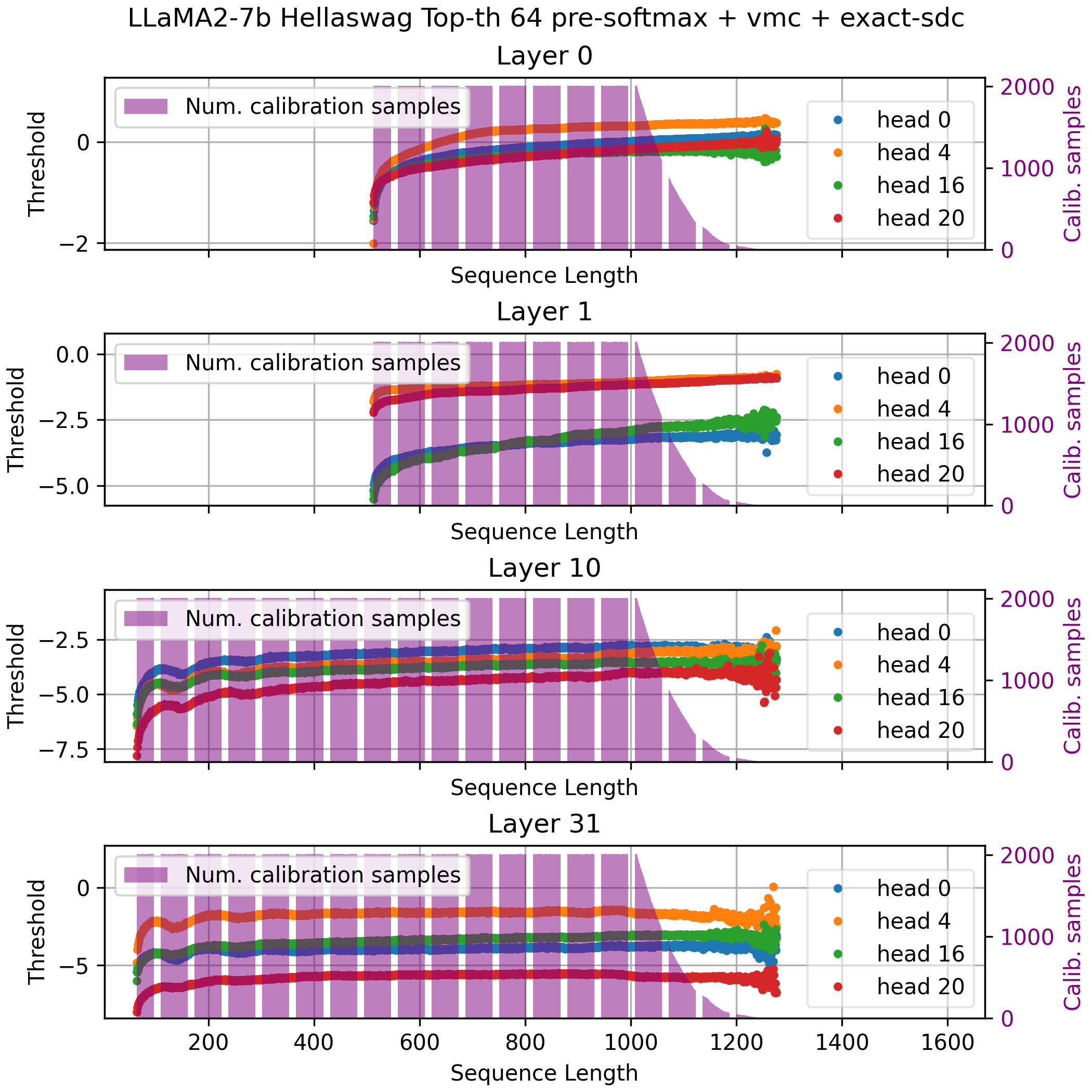}
    \caption{Pre-softmax}
    \label{fig:appendix:th_llama:pre}
\end{subfigure}
\hfill
\begin{subfigure}{0.46\textwidth}
    \includegraphics[width=\textwidth]{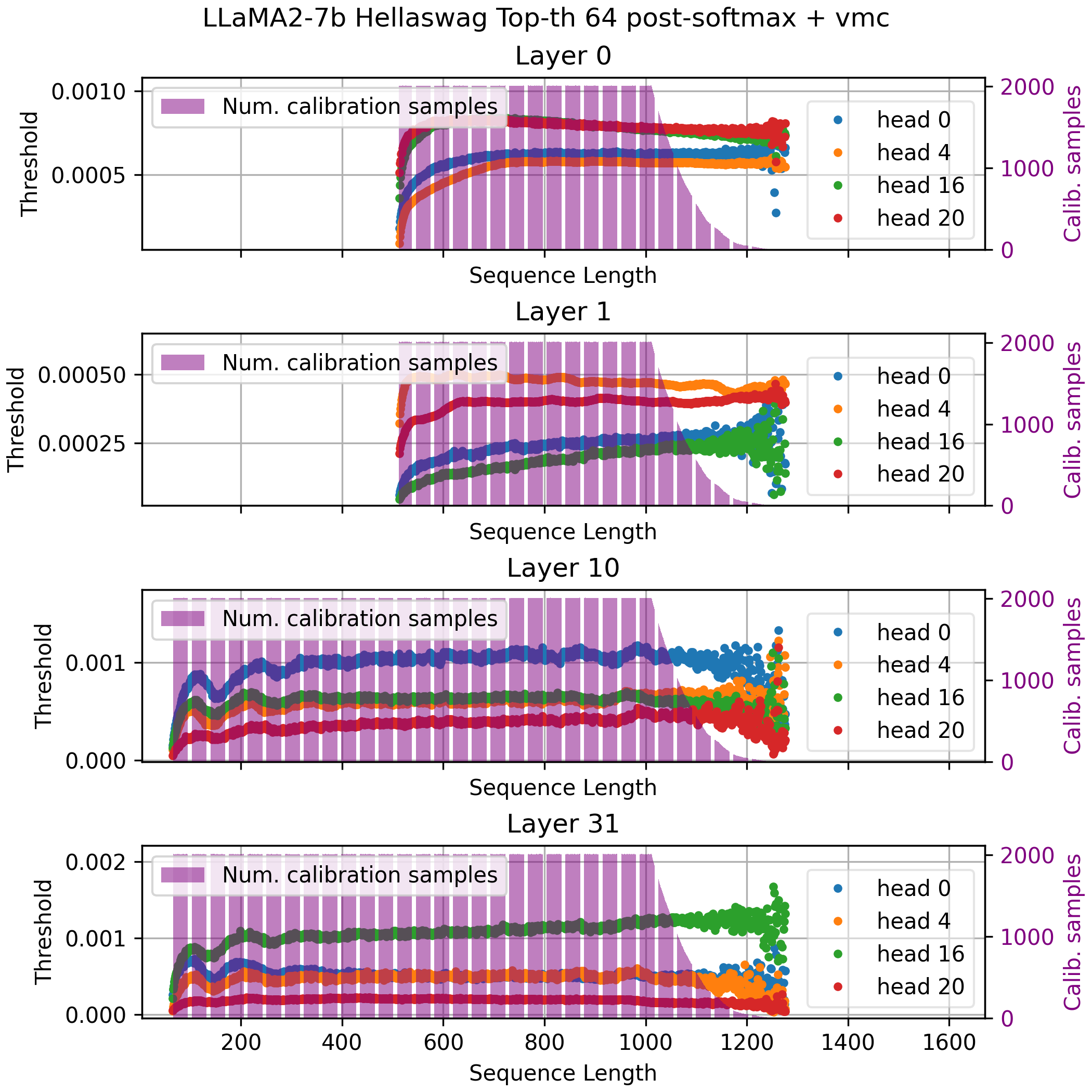}
    \caption{Post-softmax}
    \label{fig:appendix:th_llama:post}
\end{subfigure}
        
\caption{\textbf{Threshold values as a function of a sequence length in LLaMA2-7b}. Left Column: pre-softmax, Right Column: post-softmax. Scatter plots: final calibrated threshold values. Rows top to bottom: layer 0,1,10,31. Purple columns: number of calibration inputs that contained the respective sequence length. Layers 0 and 1 were calibrated for $k=512$, and layers 10 and 31 were calibrated for $k=64$.}
\label{fig:appendix:th_llama}
\end{figure}

%
In this appendix section, we show more calibrated threshold values (as a function of the sequence length) in more layers and heads. ~\cref{fig:appendix:th_llama} visualizes the thresholds as a function of a sequence length that was calibrated for The LLaMA2-7b model for the \hell dataset. Different heads have different threshold values, hence the importance of per-head individual calibrations
%
%

%
\section{Multi-k Cumulative Calibration}\label{appendix:multik_cumulitive_calibration}
\begin{figure}[ht]
    \centering
    \includegraphics[width=0.36\textwidth]{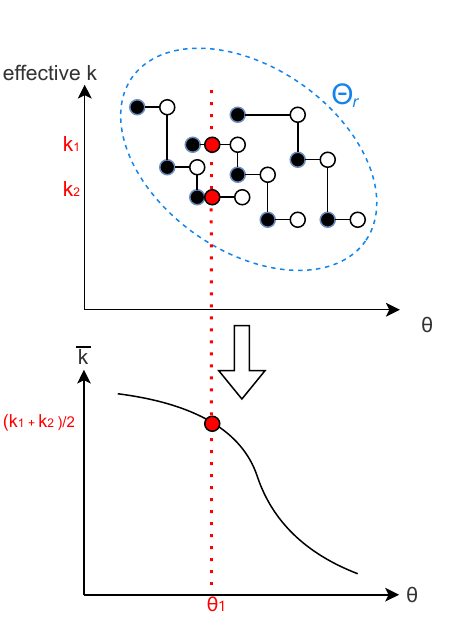}
    \caption{\textbf{MergeIntervals($\Theta_r$)} subroutine is merging the set $\Theta_r$ of interval sequences by averaging effective $k$ of all overlapping intervals across all interval sequences in the set $\Theta_r$}
    \label{fig:appendix:merge_intervals}
\end{figure}

\begin{figure}[h!]
\centering
\begin{subfigure}{0.98\textwidth}
    \includegraphics[width=\textwidth]{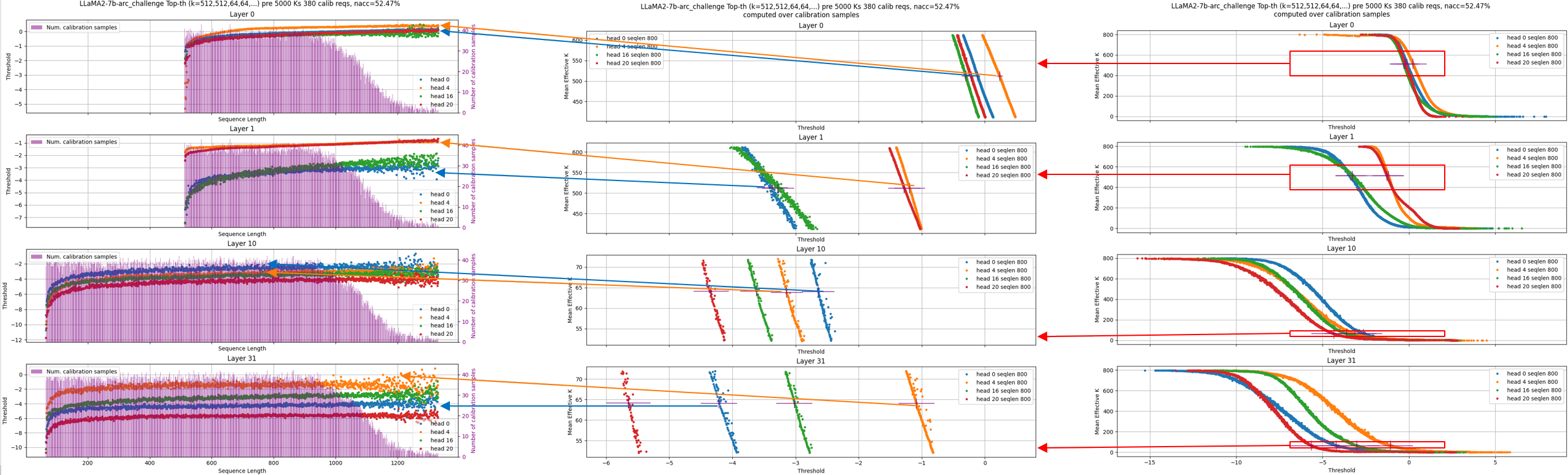}
    \caption{Pre-softmax}
    \label{fig:appendix:mkc_thresholds_visualized:pre}
\end{subfigure}
\begin{subfigure}{0.98\textwidth}
    \includegraphics[width=\textwidth]{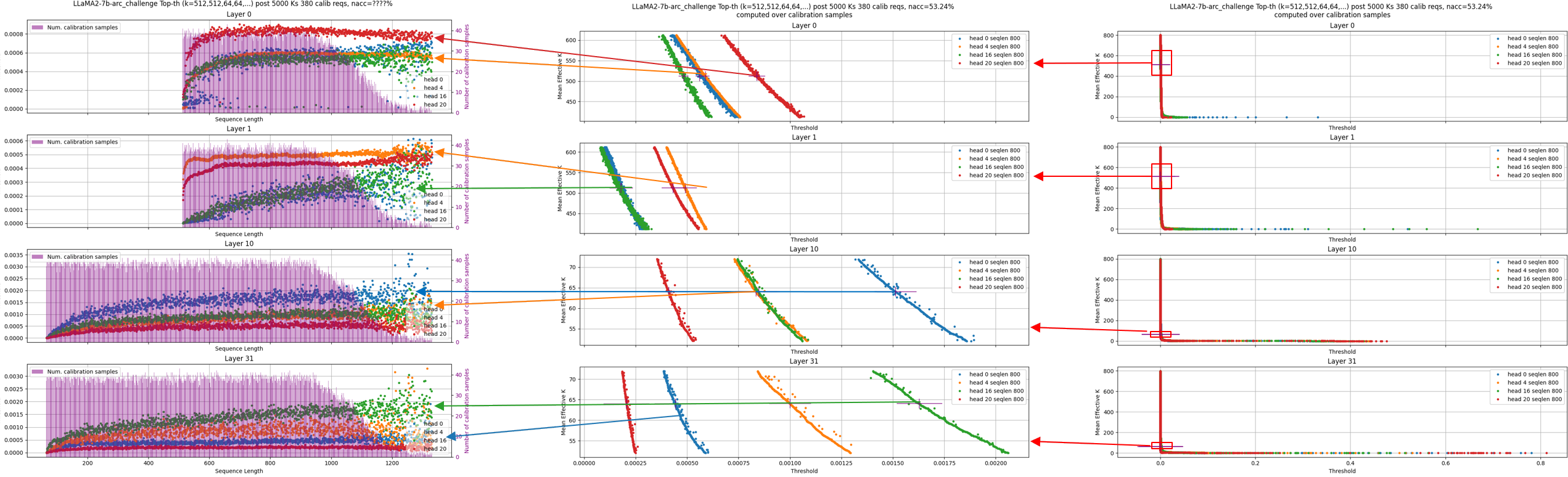}
    \caption{Post-softmax}
    \label{fig:appendix:mkc_thresholds_visualized:post}
\end{subfigure}
\caption{MKC calibration of LLaMA2-7b on \hell dataset. Right: the $\bar{k}_r(\theta)$ function as obtained from~\cref{alg:mkc} by merging the intervals for sequence length $r=800$. Middle: zoom into a range of interest of $\bar{k}_r^{-1}(k)$ around $k=64$, the selected threshold is marked on using $\theta_r(64)=\bar{k}_r^{-1}(64)$, Left: thresholds obtained from MKC for all sequence lengths (also other than $r=800$.}
\label{fig:appendix:mkc_thresholds_visualized}
\end{figure}

%
\cref{alg:calibrate}, which we presented in the paper, calibrates thresholds of a single attention head for a single target $k$ value. We propose to generalize it to calibrate all thresholds for a broader range of target $k$ and to do it in one pass over the calibration samples. We call such a calibration procedure ``Multi-k Cumulative'' (MKC).
The goal of MKC calibration is to construct a multi-$k$ threshold function $\theta(k):\mathbb{N}\rightarrow\mathbb{R}$ for every (layer $l$, head $h$, attention row id $r$). The user will be able to query such a function using their $k$ of choice and obtain the threshold needed to accommodate this $k$. Such an ability allows a very flexible control over the sparsification of the model, which is a highly desirable tool for LLM inference service that should be able to dynamically trade some accuracy for speedup.
\paragraph{MKC algorithm} ~\cref{alg:mkc} describes how a threshold function $\theta(k)$ can be calibrated. First, for every given calibration sample, let $v$ represent the $r^{th}$ attention row. The $v$ undergoes a sorting and then is treated as a sequence of half-open intervals. For each interval, we treat its smaller endpoint as its threshold, and we also associate with it an effective $k$ (effective $k$ of a threshold w.r.t the vector $v$ is the number of elements that are greater than the threshold). For example, on line 2 we represent by $\langle r-1,[v_1,v_2)\rangle$ the interval between $v_1$ (inclusive) and $v_2$ (exclusive) that corresponds to the threshold $v_1$ and an effective $k$ of $r-1$. 
Second, per-calibration-sample interval sequences collected in the set $\Theta_r$ are merged to represent the function $\bar{k}(\theta):\mathbb{R}\rightarrow\mathbb{R}$ which maps each interval to a threshold and to an average effective $k$ achievable by this threshold on the calibration set. This merging is done by the \textbf{MergeIntervals($\Theta_r$)} subroutine, which computes the following:
\begin{align}
    \bar{k}_r(\theta) = \frac{1}{|\Theta_{r,\theta}|}\sum t_0 ,\forall\theta
\end{align}
where we define $\Theta_{r,\theta}=\{t | \theta\in t_1 \wedge t\in seq \wedge seq\in\Theta_r\}$ as a subset of intervals that include the $\theta$ in the interval.
Note that the resulting average effective $k$ might be fractional due to the averaging. Finally, the desired $\theta(k)$ is given by the inverse of $\bar{k}(\theta)$, at points where the function $\bar{k}(\theta)$ maps to a natural value. The full implementation of the MergeIntervals routine is available in the supplied source code, and we show its visualization on~\cref{fig:appendix:merge_intervals}.
\begin{algorithm}[h!]
    \caption{MKC($\calibrationSet$) - 1 head calibration of thresholds for all possible $k$}
    \label{alg:mkc}
    \begin{algorithmic}[1]
        \REQUIRE \( \calibrationSet \) (Calibration set of inputs)
        \STATE $\Theta_{r}=\emptyset, \forall r $ \COMMENT{empty sets of observed interval sequences}
        \FOR{\( X \in \calibrationSet \)}
            \IF{is\_prefill($X$)} 
                \STATE $\attnMatPre = Q(X)K^T(X)$
                \STATE $\seqlen = $NumRows$(\attnMatPre)$
                \FOR{\( r = 1 \) to \( \seqlen - 1 \)}
                    \STATE $\bm{v} = $Sort$(\attnMatPre_r)$
                    \STATE $\Theta_{\seqlen}$ = $\Theta_{\seqlen}\cup\{\langle r-1,[v_0,v_1)\rangle,\langle r-2,[v_1,v_2)\rangle,\ldots,\langle 1,[v_{r-2},v_{r-1})\rangle\}$           
                \ENDFOR
            \ELSE[generative decoding, $X\in\mathbb{R}^\headdim$]
                \STATE $\attnVecPre = Q(X)K^T(X)$
                \STATE $\seqlen = $Length$(\attnVecPre)$            
                \STATE $\bm{v} = $Sort$(\attnVecPre)$
                \STATE $\Theta_{\seqlen}$ = $\Theta_{\seqlen}\cup\{\langle n-1,[v_0,v_1)\rangle,\langle n-2,[v_1,v_2)\rangle,\ldots,\langle 1,[v_{n-2},v_{n-1})\rangle\}$           
            \ENDIF
        \ENDFOR
        \STATE $\bar{k}_r(\theta) = $MergeIntervals$(\Theta_r), \forall r$
        \STATE \textbf{Return} $\theta_r(k) = \bar{k}_r^{-1}(k), \forall r$
    \end{algorithmic}
\end{algorithm}
~\cref{fig:appendix:mkc_thresholds_visualized} illustrates the distribution of merged intervals ($\bar{k}(\theta)$) for selected attention layers and heads in the LLaMA2-7B model, following calibration via MKC on the \hell dataset.
%
%
%

%
It is worth mentioning that the main downside of the multi-k calibration is that it is very time and memory-consuming to perform all these interval mergings individually for every layer, head, and attention row id (i.e., sequence length). Speeding up the MergeIntervals routine, using reasonable approximations (such as subsampling) is an interesting research direction. The second slight downside is that it does not allow to apply top$_k$ at calibration, as we used to apply in~\cref{alg:calibrate}, since during MKC there is a multitude of potential $k$ parameters we are targeting. 

\clearpage
\section{V-Mean Compensation}\label{appendix:vmc}
In this section, we provide a formal proof for our $\bm{V}$-Mean Compensation (VMC)  serving as a good approximation of the sparsified $\attnVecSparsePost\Tilde{\bm{V}}$ product to the full $\mathbf{\attnVecPost}\bm{V}$ product.
\begin{lemma}
 Let $\attnVecPost\in[0,1]^\seqlen$ denote the post-softmax attention vector, and let $\attnVecSparsePost\in[0,1]^{\Tilde{k}}$ denote its thresholded variant, containing the $\Tilde{k}$ selected values of $\attnVecPost$, and let $\selids\in\left\{0,\dots, n-1\right\}^{\Tilde{k}}$ denote the indices of $\attnVecPost$ that surpassed the threshold such that $\forall i\in\{0,\dots, \Tilde{k}-1\}: \attnVecSparsePost_i=\attnVecPost_{\selids_i}$. Let $\bm{V}\in\mathbb{R}^{n\times d}$ be the value matrix and let $\Tilde{\bm{V}}\in\mathbb{R}^{\Tilde{k}\times d}$ be  consisting only of the selected $\bm{V}$ rows such that 
 $\forall i\in\{0,\dots, \Tilde{k}-1\}: \Tilde{\bm{V}}_i=\bm{V}_{\selids_i}$.
 We claim that in the expectation, the full product $\bm{p}=\attnVecPost \bm{V}\in\mathbb{R}^\headdim$ is equal to the thresholded attention plus the residual probability mass $\beta$ (\ref{eqn:vmc:beta}) multiplied by the mean $\bm{V}$ row $\bm\mu$ (\ref{eqn:vmc:mu}). Namely,
\begin{align}
    \forall 0\le j<\headdim: E[\bm{p}_j] = (\bm{\Tilde{s}}\Tilde{\bm{V}})_j + \beta\bm{\mu}_j
\end{align}

\end{lemma}
\begin{proof}
\begin{equation}\label{eqn:vmc:proof}
    \begin{split}
    E[\bm{p}_j] &= E\Big[\sum_{i=0}^{\seqlen-1}s_i\bm{V}_{ij}\Big] \\
    &= E\Big[\sum_{i\in\selids}s_i \bm{V}_{ij}\Big] + E\Big[\sum_{i\in\notselids}s_i V_{ij}\Big] \\
    &= (\bm{\Tilde{s}}\Tilde{\bm{V}})_j + \sum_{i\in\notselids}E[s_i \bm{V}_{ij}] \\ 
    &\underset{s\ind \bm{V}_j}{=} (\bm{\Tilde{s}}\Tilde{\bm{V}})_j + \sum_{i\in\notselids}E[s_i]E[\bm{V}_{ij}] \\ &\underset{s,\bm{V}_j\text{uniform}}{=} (\bm{\Tilde{s}}\Tilde{\bm{V}})_j + \sum_{i\in\notselids}\frac{\beta}{\seqlen-\numKept}\bm{\mu}_j \\
    &= (\bm{\Tilde{s}}\Tilde{\bm{V}})_j + (\seqlen-\numKept)\frac{\beta}{\seqlen-\numKept}\bm{\mu}_j \\
    &= (\bm{\Tilde{s}}\Tilde{\bm{V}})_j + \beta\bm{\mu}_j
    \end{split}
\end{equation}
\end{proof}
Under the following assumptions:

\begin{enumerate}
    \item $s\ind \bm{V}_j$, that is the attention vector $\attnVecPost$ is statistically independent on the elements in the columns of matrix $\bm{V}$. They are conditionally independent given the input $X$ from which they were originally computed via $\bm{V}=\bm{XW_V}$.
    \item  The distribution of $s_i, \forall i\in\notselids$ within the long tail of the non-selected indices is close to uniform, and hence we can approximate its expectation by an average.
    \item The expectation of $V_{ij}$ can be approximated by its average.
\end{enumerate}

\section{Evaluation statistics}\label{appendix:evaluation_statistics}
\begin{figure}
\centering
\begin{subfigure}{0.32\textwidth}
    \includegraphics[width=\textwidth]{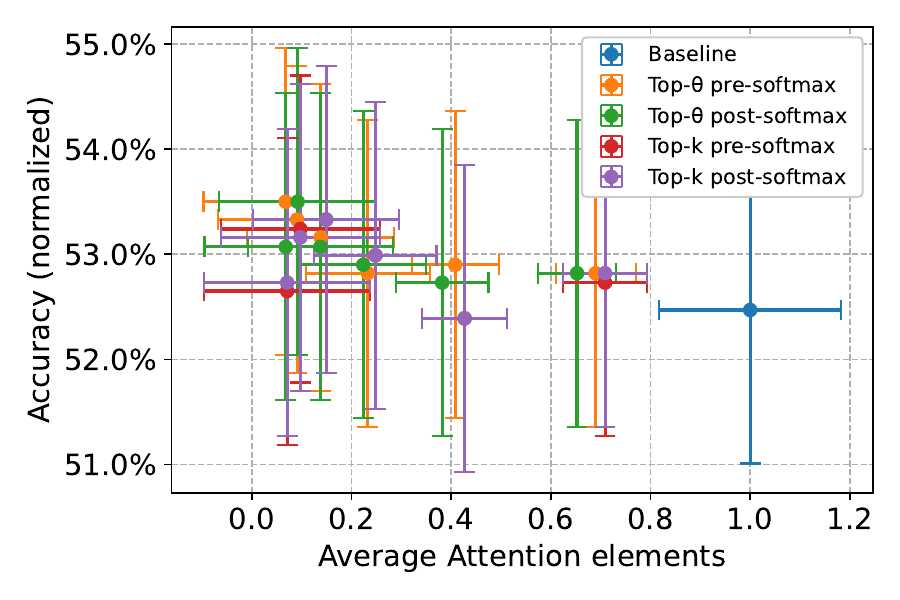}
    \caption{LLaMA2-7b \arcc}
    \label{fig:appendix:evaluations:acc_kept_attn:llama2_arc_challenge}
\end{subfigure}
\hfill
\begin{subfigure}{0.32\textwidth}
    \includegraphics[width=\textwidth]{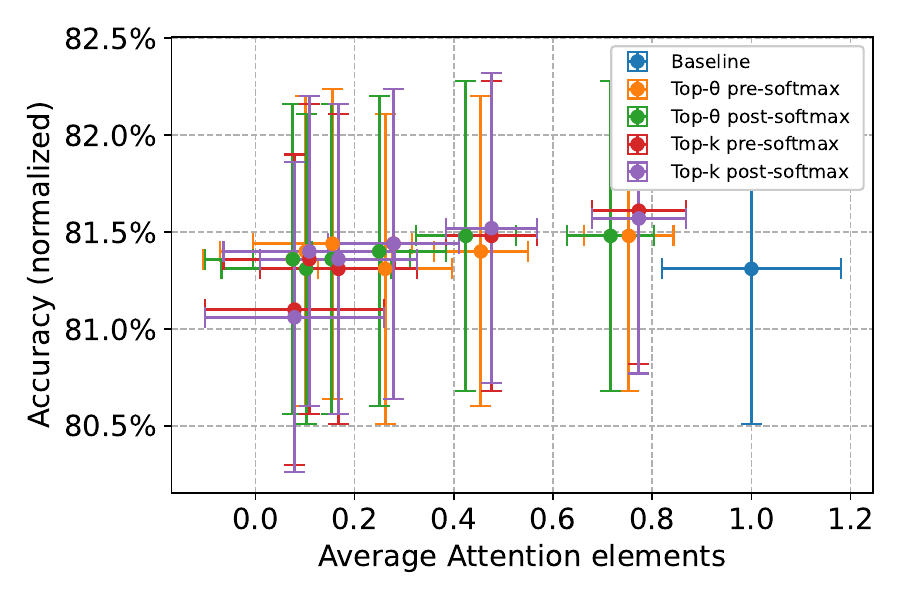}
    \caption{LLaMA2-7b \arce}
    \label{fig:appendix:evaluations:acc_kept_attn:llama2_arc_easy}
\end{subfigure}
\hfill
\begin{subfigure}{0.32\textwidth}
    \includegraphics[width=\textwidth]{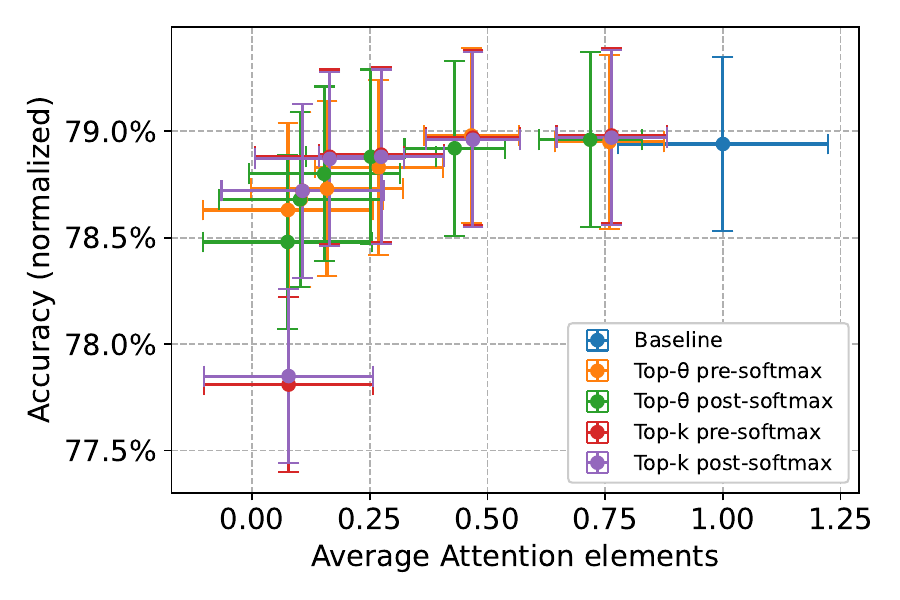}
    \caption{LLaMA2-7B \hell }
    \label{fig:appendix:evaluations:acc_kept_attn:llama2_hellaswag}
\end{subfigure}

\begin{subfigure}{0.32\textwidth}
    \includegraphics[width=\textwidth]{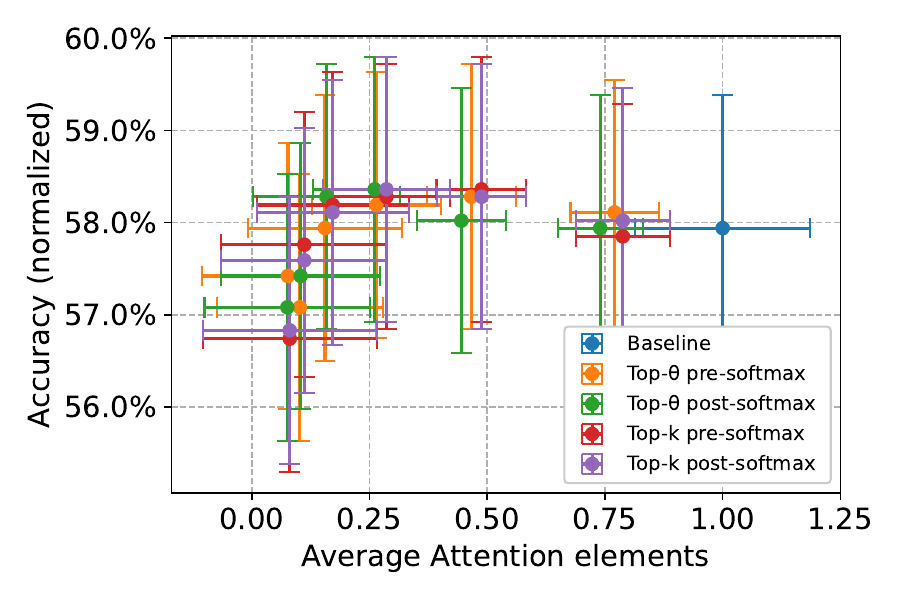}
    \caption{LLaMA3-8B \arcc}
    \label{fig:appendix:evaluations:acc_kept_attn:llama3_arc_challenge}
\end{subfigure}
\hfill
\begin{subfigure}{0.32\textwidth}
    \includegraphics[width=\textwidth]{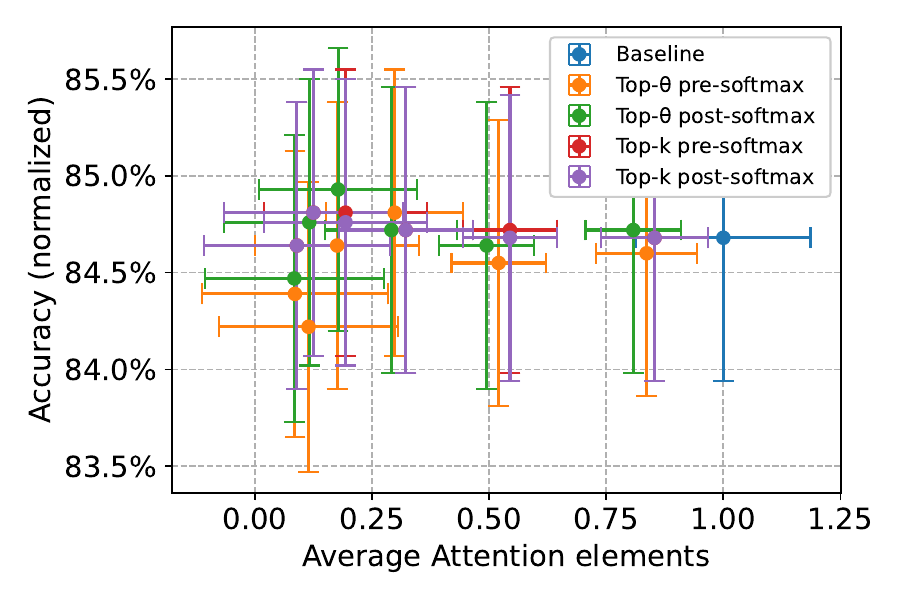}
    \caption{LLaMA3-8B \arce}
    \label{fig:appendix:evaluations:acc_kept_attn:llama3_arc_easy}
\end{subfigure}
\hfill
\begin{subfigure}{0.32\textwidth}
    \includegraphics[width=\textwidth]{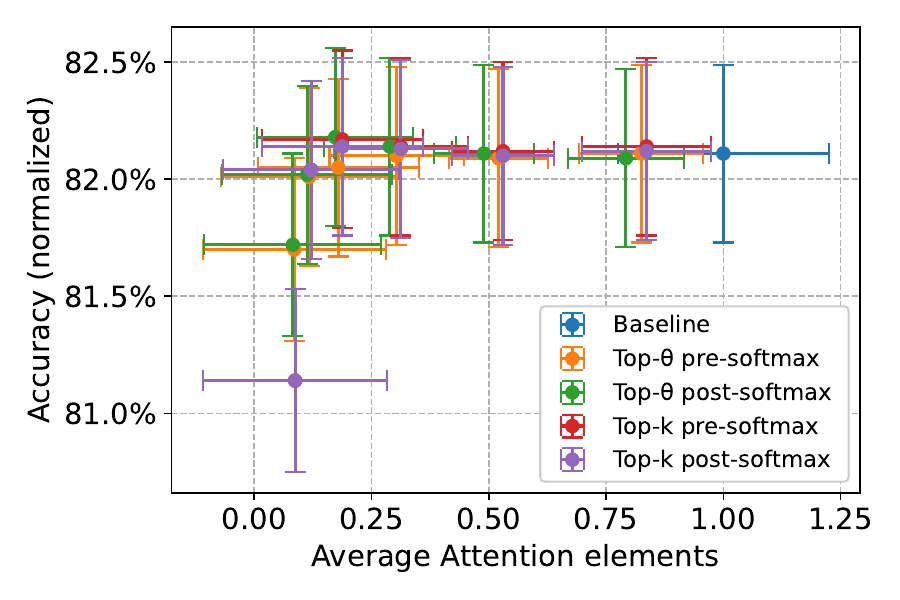}
    \caption{LLaMA3-8B \hell }
    \label{fig:appendix:evaluations:acc_kept_attn:llama3_hellaswag}
\end{subfigure}

\begin{subfigure}{0.32\textwidth}
    \includegraphics[width=\textwidth]{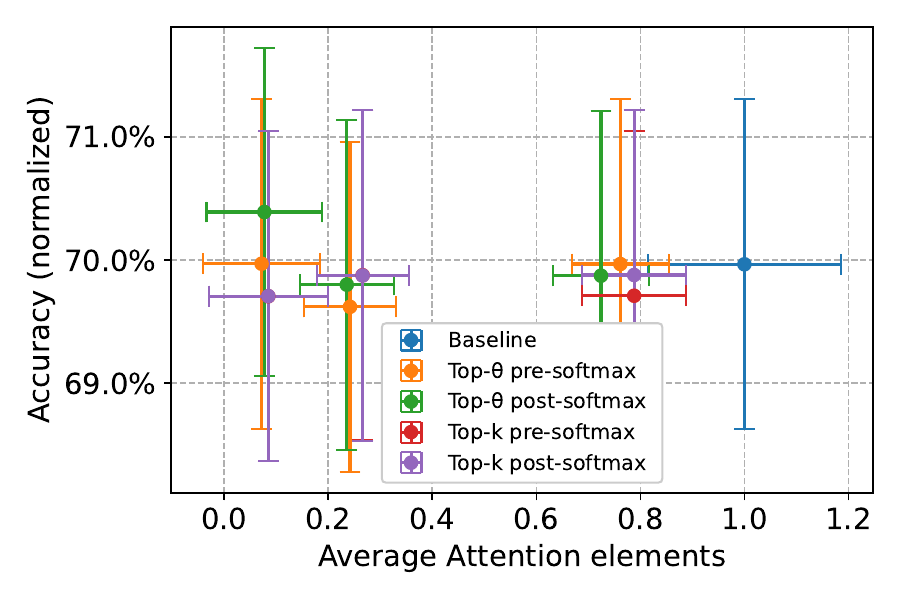}
    \caption{LLaMA3-70B \arcc}
    \label{fig:appendix:evaluations:acc_kept_attn:llama3_70_arc_challenge}
\end{subfigure}
\hfill
\begin{subfigure}{0.32\textwidth}
    \includegraphics[width=\textwidth]{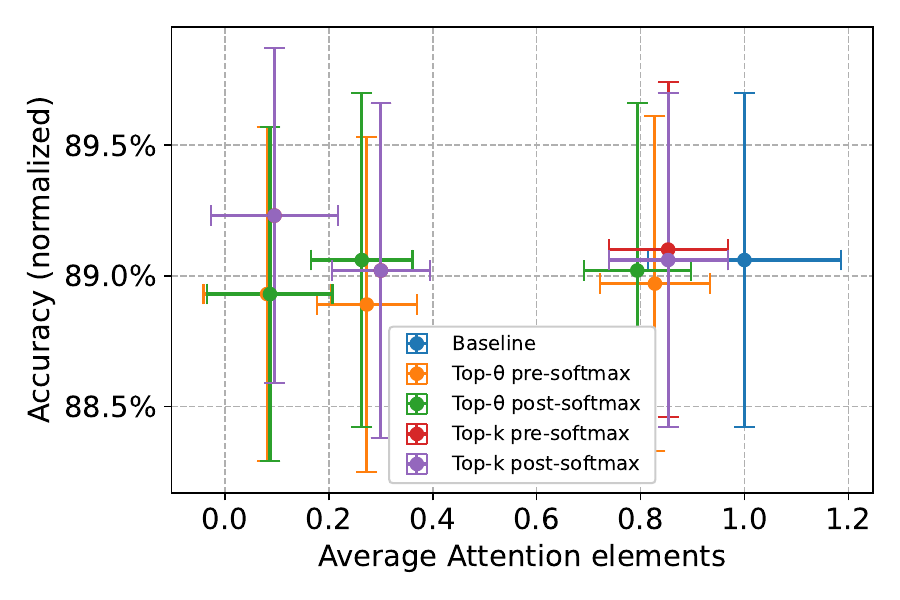}
    \caption{LLaMA3-70B \arce}
    \label{fig:appendix:evaluations:acc_kept_attn:llama3_70_arc_easy}
\end{subfigure}
\hfill
\begin{subfigure}{0.32\textwidth}
    \includegraphics[width=\textwidth]{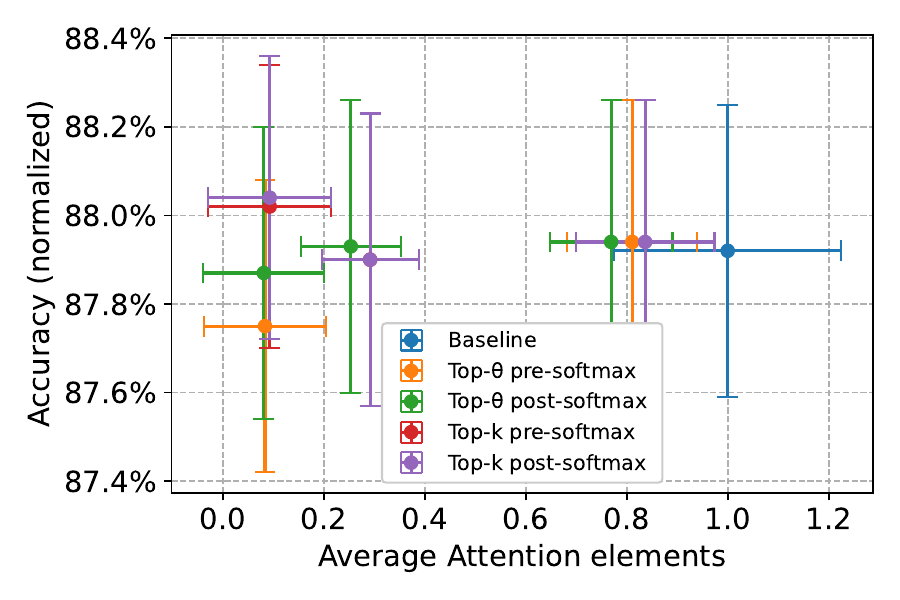}
    \caption{LLaMA3-70B \hell }
    \label{fig:appendix:evaluations:acc_kept_attn:llama3_70_hellaswag}
\end{subfigure}

\caption{\textbf{Prefill-based tasks} - Tradeoff between model accuracy averaged across test samples (y-axis), and the portion of kept attention elements per attention head (x-axis). All post-softmax \topk and \topth employ VMC, and all pre-softmax variants employ both VMC and exact SDC. Using the VMC and the SDC compensations achieves little if any accuracy degradation while achieving up to 10\texttimes\ reduction in the attention elements.}
\label{fig:appendix:evaluations:acc_kept_attn}
\end{figure}

\begin{figure}
\centering
\begin{subfigure}{0.43\textwidth}
    \includegraphics[width=\textwidth]{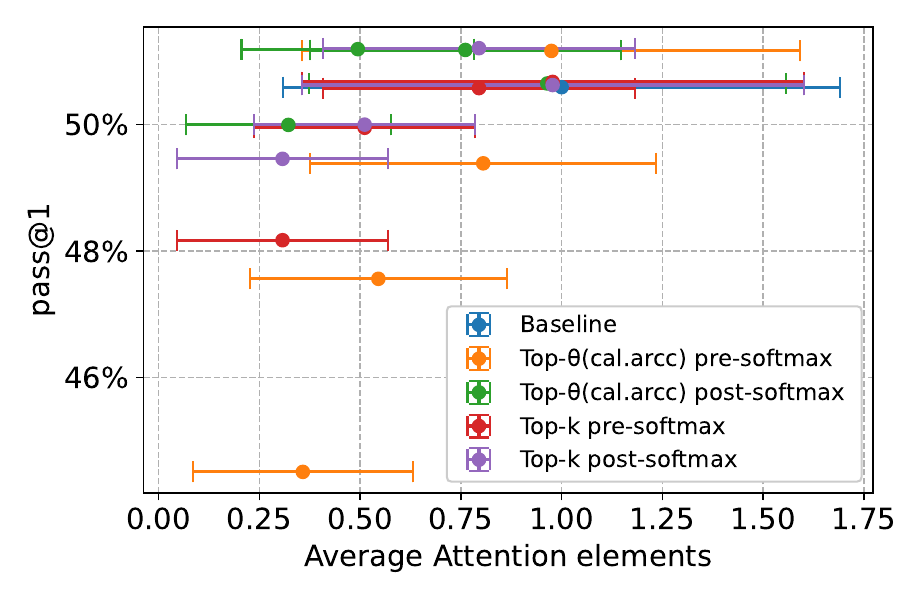}
    \caption{LLaMA3-8B-Instruct HumanEval}
    \label{fig:appendix:evaluations_generative:acc_kept_attn:llama3_8b_humaneval}
\end{subfigure}
\hfill
\begin{subfigure}{0.43\textwidth}
    \includegraphics[width=\textwidth]{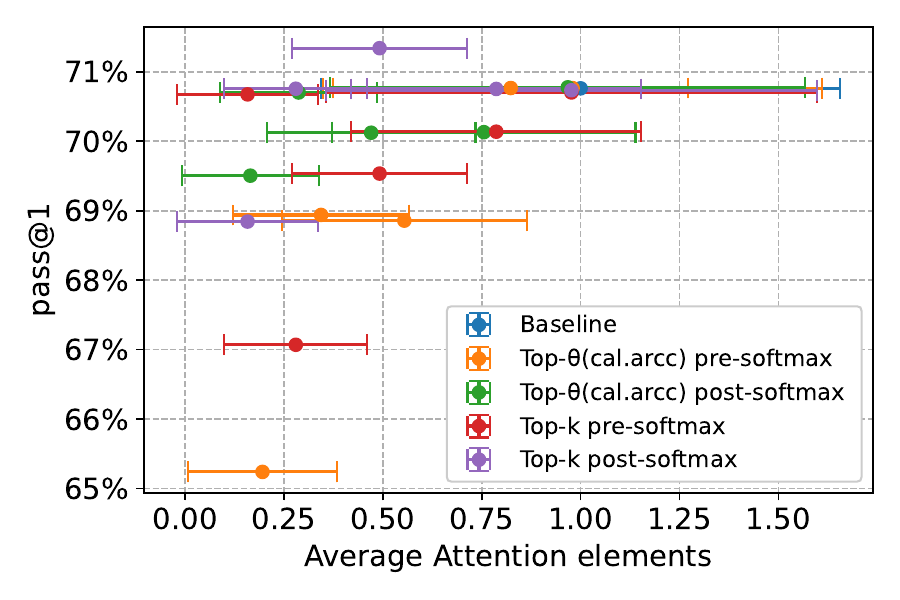}
    \caption{LLaMA3-70B-Instruct HumanEval}
    \label{fig:appendix:evaluations_generative:acc_kept_attn:llama3_70b_humaneval}
\end{subfigure}
\begin{subfigure}{0.46\textwidth}
    \includegraphics[width=\textwidth]{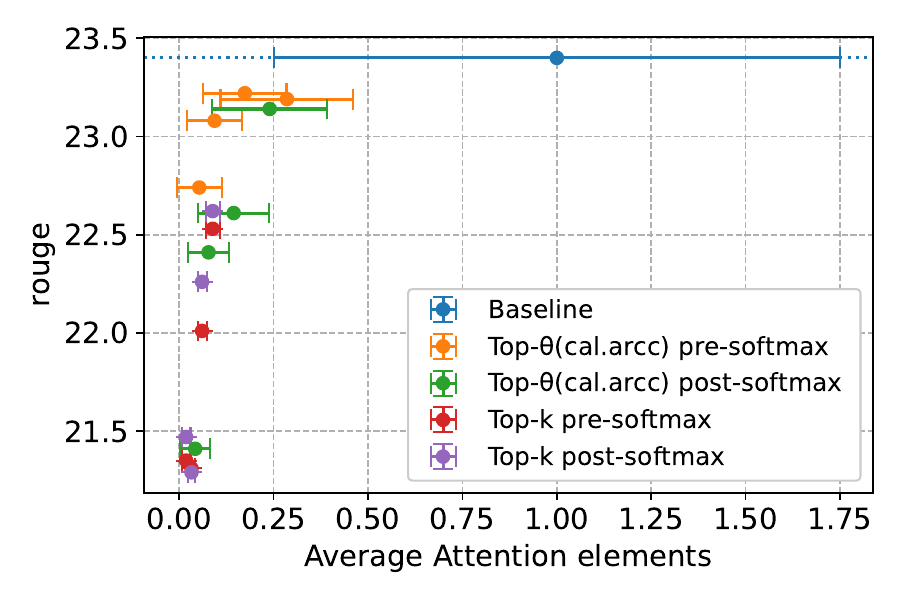}
    \caption{LLaMA3.1-8B-Instruct \qmsum}
    \label{fig:appendix:evaluations_generative:acc_kept_attn:llama3.1_8B_qmsum}
\end{subfigure}
\hfill
\begin{subfigure}{0.46\textwidth}
    \includegraphics[width=\textwidth]{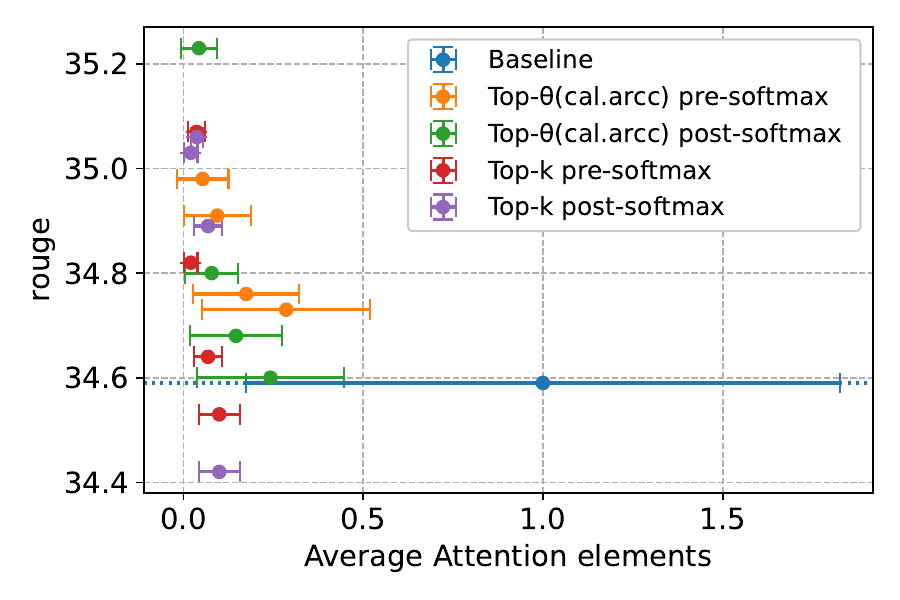}
    \caption{LLaMA3.1-8B-Instruct \govreport}
    \label{fig:appendix:evaluations_generative:acc_kept_attn:llama3.1_8B_gov_report}
\end{subfigure}
\caption{\textbf{Generative Tasks - during prefill} - Tradeoff between model accuracy averaged across test samples (y-axis), and the portion of required attention elements per head (x-axis). The \topth variants employ threshold calibrated on \arcc dataset. All post-softmax \topk and \topth employ VMC, all pre-softmax variants employ both VMC and exact SDC}
\label{fig:appendix:evaluations_generative:acc_kept_attn}
\end{figure}

\begin{figure}
\centering
\begin{subfigure}{0.43\textwidth}
    \includegraphics[width=\textwidth]{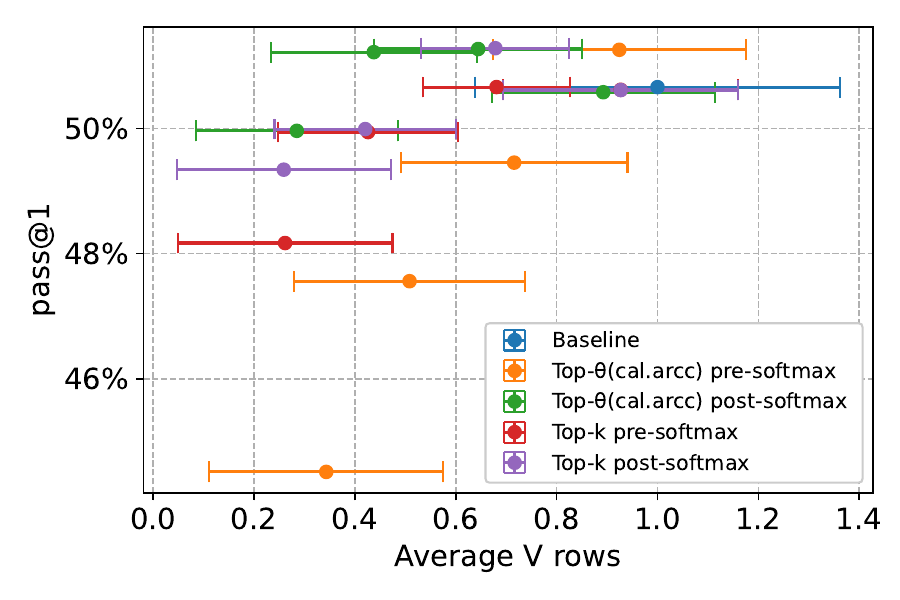}
    \caption{LLaMA3-8B-Instruct \humaneval}
    \label{fig:appendix:evaluations_generative:acc_kept_vrow:llama3_8b_humaneval}
\end{subfigure}
\hfill
\begin{subfigure}{0.43\textwidth}
    \includegraphics[width=\textwidth]{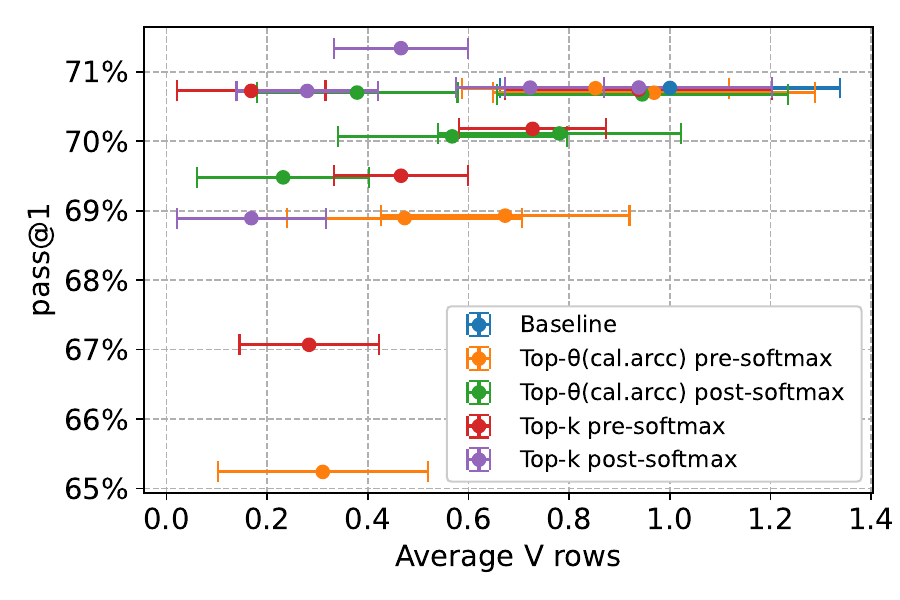}
    \caption{LLaMA3-70B-Instruct \humaneval}
    \label{fig:appendix:evaluations_generative:acc_kept_vrow:llama3_70b_humaneval}
\end{subfigure}
\hfill
\begin{subfigure}{0.46\textwidth}
    \includegraphics[width=\textwidth]{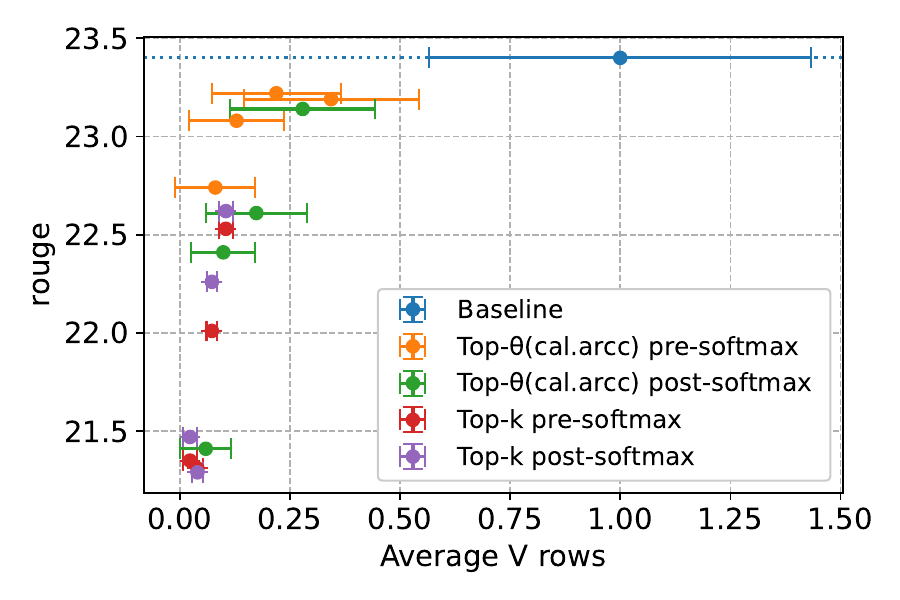}
    \caption{LLaMA3.1-8B-Instruct \qmsum}
    \label{fig:appendix:evaluations_generative:acc_kept_vrow:llama3.1_8B_qmsum}
\end{subfigure}
\hfill
\begin{subfigure}{0.46\textwidth}
    \includegraphics[width=\textwidth]{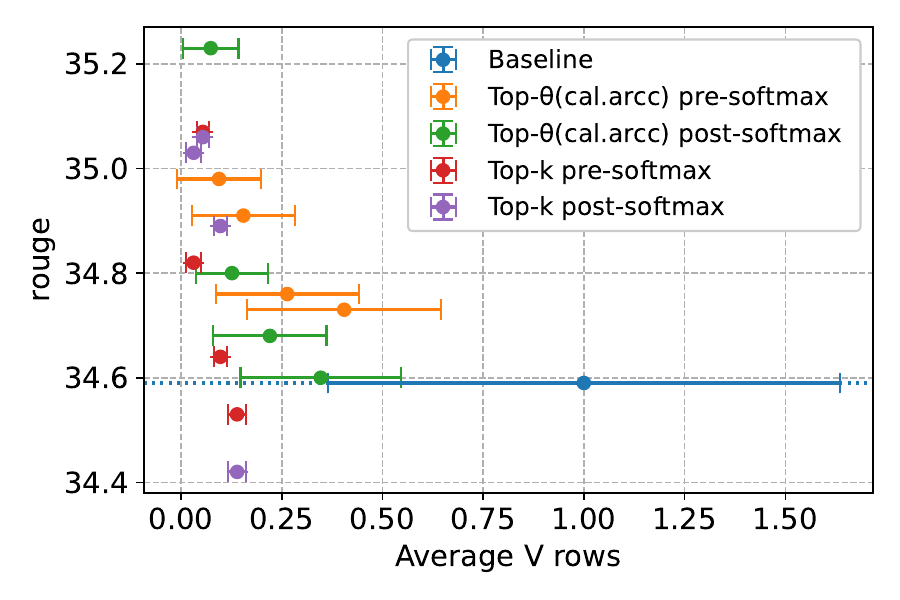}
    \caption{LLaMA3.1-8B-Instruct \govreport}
    \label{fig:appendix:evaluations_generative:acc_kept_vrow:llama3.1_8B_gov_report}
\end{subfigure}
\caption{\textbf{Generative Tasks - during generative decoding} -- Tradeoff between model accuracy averaged across test samples (y-axis) and the portion of required $\bm{V}$-rows per group of heads (x-axis). The \topth variants employ threshold calibrated on \arcc dataset. All post-softmax \topk and \topth employ VMC, and all pre-softmax variants employ VMC and exact SDC. 3\texttimes\ to 10\texttimes\ reduction of $\bm{V}$ rows is achieved.}
\label{fig:appendix:evaluations_generative:acc_kept_vrow}
\end{figure}

In this section, we present again the experimental results from ~\cref{sec:evaluations:overall}; however, to demonstrate statistical significance, we show the error bars. This is important since every data point is aggregated using averaging across layers, heads, and test examples, as we will describe below. Therefore standard deviation of such an averaged metric is of interest. 
\cref{fig:appendix:evaluations:acc_kept_attn} focus on Q\& tasks, showing the tradeoff between the model's accuracy (y-axis) and the number of attention elements as selected by the \topk or \topth (x-axis). 
The standard deviation in the accuracy was provided by the LM-Eval evaluation harness, as a part of the standardized evaluation procedure. %
The average ratio presented on the x-axis and its standard deviation were computed over the following population: number of test samples $\times$ number of model layers $\times$ number of attention heads. To present the ratios, we first compute the absolute average and absolute standard deviation of the total count of attention elements and in the attention matrix, second - we normalize both the standard deviation and the average by the average number of attention elements when no sparsification took place.
\cref{fig:appendix:evaluations_generative:acc_kept_attn,fig:appendix:evaluations_generative:acc_kept_vrow} show the evaluation on \humaneval dataset. ~\cref{fig:appendix:evaluations_generative:acc_kept_attn} is similar to the Q\&A plots as it shows how the reduction in the number of attention elements during \textit{prefill phase} impacts the accuracy score (pass@1). \cref{fig:appendix:evaluations_generative:acc_kept_vrow} focuses on the generative \textit{decoding phase} and shows how the number of the needed $\bm{V}$ rows is affected by the elements that were selected in every row independently, which introduces the effect of GQA - since the indices of the selected elements are united across the heads in the group. The average and the standard deviation in this plot are taken across the following population of samples: number of test tasks $\times$ number of autoregressive forward passes $\times$ number of model layers $\times$ number of attention heads $\times$.
%
%

%
%

%
%

\section{Pre- vs. post-softmax sparsification}\label{appendix:pre-vs-post-softmax}
This section provides extended experiment plots comparing pure pre-softmax and post-softmax accuracy on Q\&A tasks. In all of them the post softmax consistently achieves higher scores.
\begin{figure}[t]
    \centering
    \includegraphics[width=0.46\textwidth]{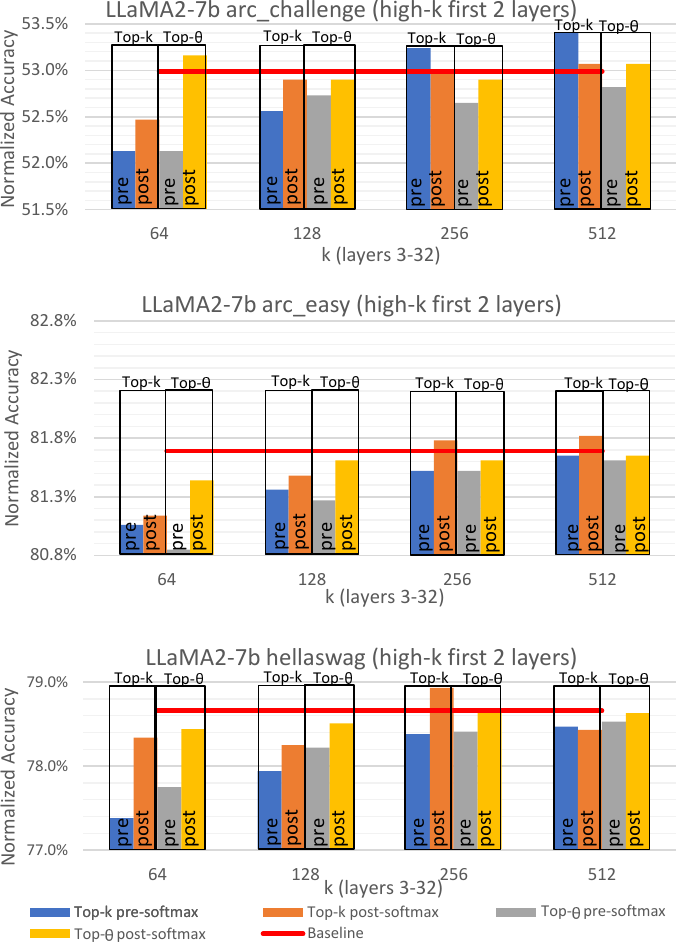}
    \caption{Comparison of Pre- and post-softmax thresholding}
    \label{fig:appendix:pre-post}
\end{figure}

%
\clearpage
\section{Thresholding different layers}\label{appendix:layers}
\begin{figure}[h!]
    \centering 
    \includegraphics[width=0.98\textwidth]{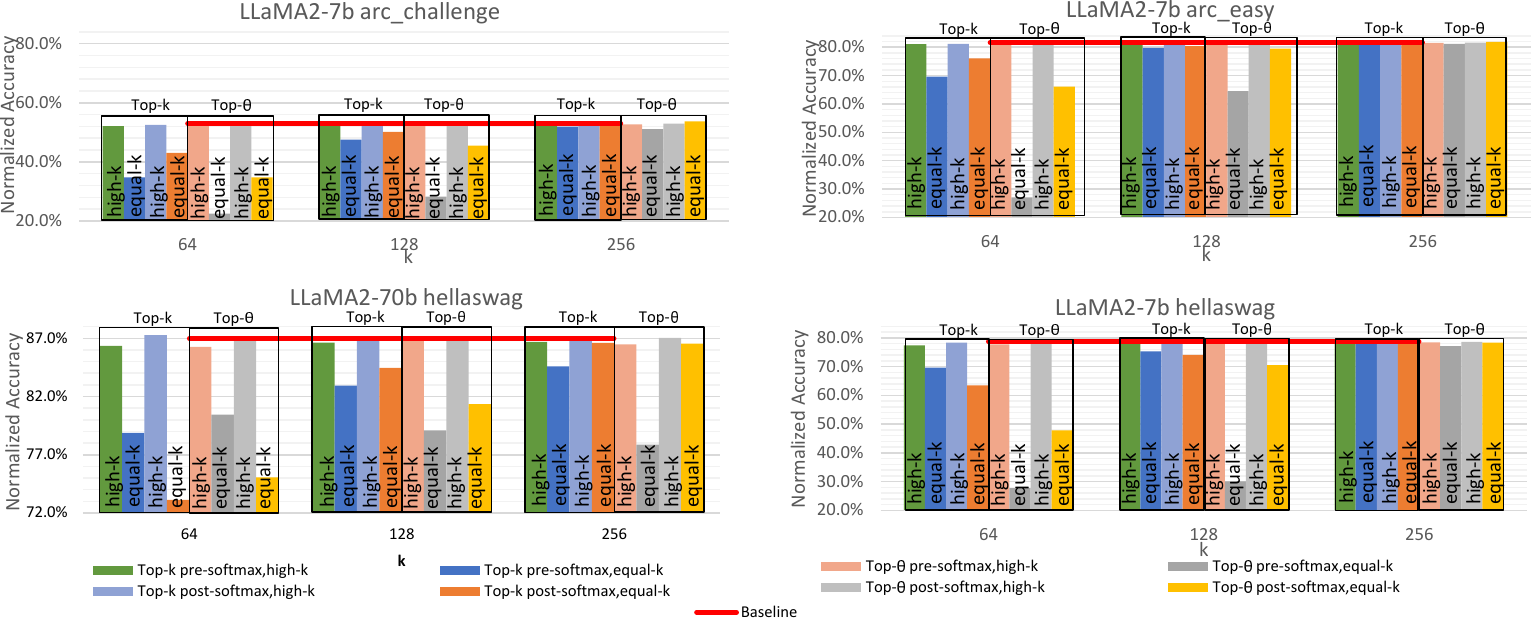} 
    \caption{The positive impact of keeping first two layers dense (higher k for calibration), compared to keeping equal k in all layers}  
    \label{fig:appendix:highK} 
\end{figure}

%
In this appendix, we present more results that support the decision to use denser initial layers. That is to use higher $k$ for \topk or calibrating towards a higher $k$ in \topth. ~\cref{fig:appendix:highK} shows LLaMA2-7b model accuracy on Q\&A datasets, and LLaMA2-70b model accuracy for \hell dataset. The figure compares \topk and \topth using higher $k$ in the first 2 layers against using equal $k$ in all layers. All variants do not perform any compensations.

%
\section{Thresholding different attention rows}\label{appendix:evaluations:rows}
%
\begin{figure}[h!]
    \centering 
    \includegraphics[width=0.46\textwidth]{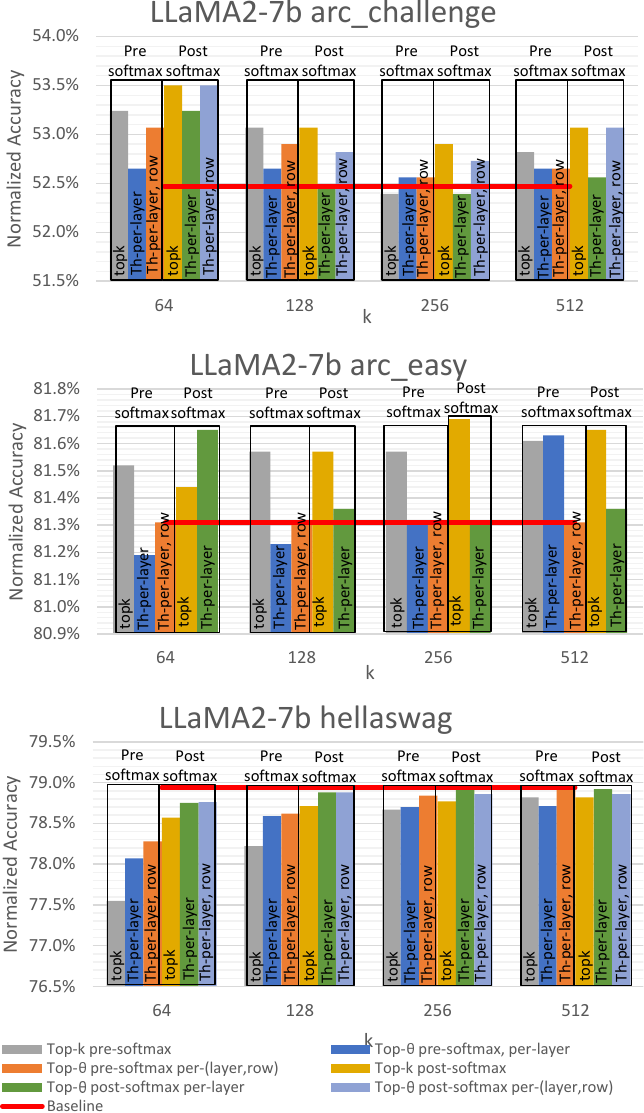} 
    \caption{Calibrating per-attention-row thresholds vs a unified threshold for all rows (sequence lengths)}  
    \label{fig:appendix:calibration_per_row} 
\end{figure}

%
\clearpage
\section{Impact of GQA}\label{appendix:evaluations:impactofgqa}
\begin{figure}[h!]
\centering
\begin{subfigure}{0.53\textwidth}
    \includegraphics[width=\textwidth]{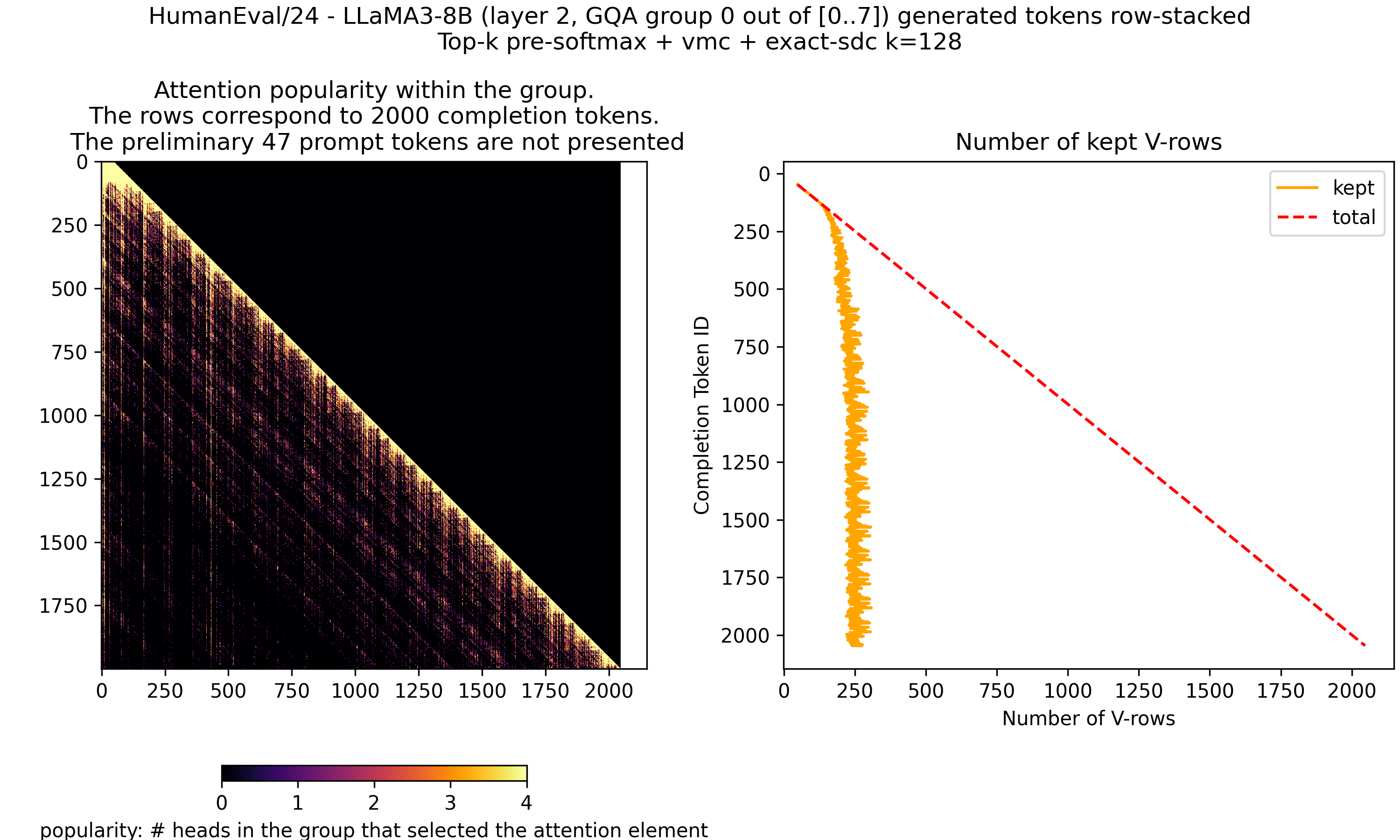}
    \caption{\topk attention with $k=128$}
    \label{fig:appendix:evaluations:gqa:topk} 
\end{subfigure}

\begin{subfigure}{0.53\textwidth}
    \includegraphics[width=\textwidth]{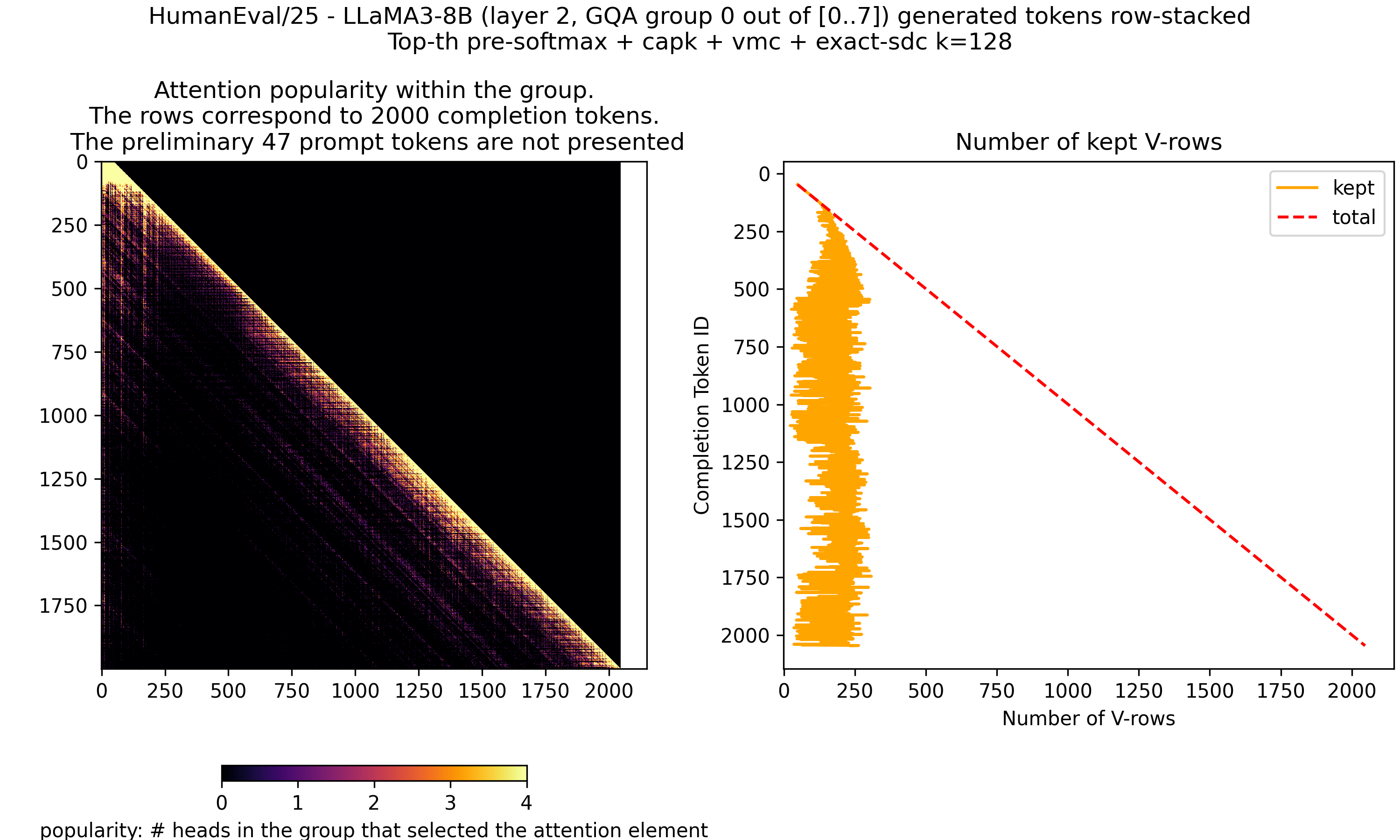}
    \caption{\topth with CAPK, and $k=128$ }
    \label{fig:appendix:evaluations:gqa:topth_capk} 
\end{subfigure}
\begin{subfigure}{0.53\textwidth}
    \includegraphics[width=\textwidth]{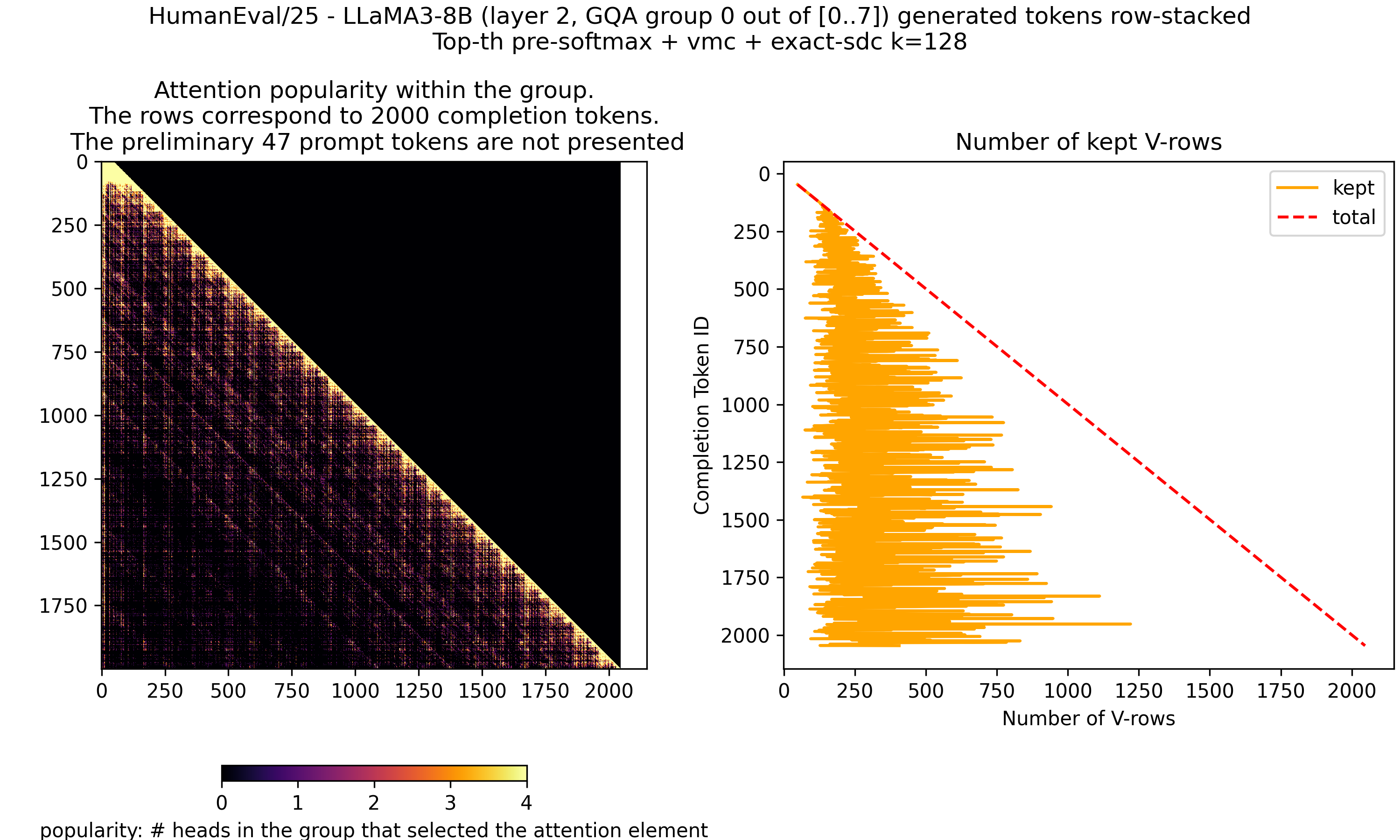}
    \caption{\topth with $k=128$ without CAPK}
    \label{fig:appendix:evaluations:gqa:topth} 
\end{subfigure}
        
\caption{\textbf{Attention popularity mask} -- LLaMA-3-8B (GQA group size$=4$), \humaneval task number 25, generative decoding iterations as rows. Left -- heat map showing how many heads had the corresponding attention element in their Top-128; on the right -- the number of $\bm{V}$-rows required to be used (1 head in the group is enough to require a $\bm{V}$ row).}
\label{fig:appendix:evaluations:gqa}
\end{figure}

%
In~\cref{fig:appendix:evaluations:gqa}, we show how the different attention sparsification approaches (\topk, \topth with and without CAPK) affect the number of required $\bm{V}$ rows. \topk (~\cref{fig:appendix:evaluations:gqa:topk}) guarantees exactly 128 selected elements per row of every head. However, the unified set of 4 heads in the group reaches only about 250, which indicates a certain agreement between the heads, which is mostly found in the recent tokens (as seen on the heatmap). We observed very similar characteristics in other heads and layers. \topth approach with capping the number of selected elements to at most 128 per head yielded degraded quality of the generated text since it mainly focused attention on the most recent tokens. Finally, the ordinary \topth in~\cref{fig:appendix:evaluations:gqa:topth}, which provided good quality results, does seem to exhibit a certain variability in the number of selected elements per group, sometimes selecting more than 128 per group.

\end{document}